\pdfoutput=1
\PassOptionsToPackage{dvipsnames}{xcolor}
\documentclass[11pt]{article}

\usepackage[final]{acl}

\usepackage{times}
\usepackage{latexsym}

\usepackage[T1]{fontenc}

\usepackage[utf8]{inputenc}

\usepackage{microtype}

\usepackage{inconsolata}
\usepackage{hyperref}
\usepackage{url}
\usepackage{kotex}

\usepackage{amsmath}
\usepackage{amssymb}
\usepackage{mathtools}
\usepackage{bbold}
\usepackage{amsthm}
\usepackage{array}
\usepackage{pgfplots}
\usepackage{tikz}
\pgfplotsset{compat=1.17}
\theoremstyle{plain}

\theoremstyle{definition}

\theoremstyle{remark}

\usepackage[utf8]{inputenc} 
\usepackage[T1]{fontenc}    
\usepackage{hyperref}       
\usepackage{url}            
\usepackage{booktabs}       
\usepackage{amsfonts}       
\usepackage{nicefrac}       
\usepackage{microtype}      
\usepackage{adjustbox} 
\usepackage{pifont}
\definecolor{darkgreen}{rgb}{0.0, 0.5, 0.0} 
\definecolor{darkred}{rgb}{0.5, 0.0, 0.0}   
\usepackage{tablefootnote}
\usepackage{graphicx}
\usepackage{subcaption}
\usepackage{bm}
\usepackage{float}
\usepackage{enumitem}

\usepackage{color, colortbl}
\definecolor{Gray}{gray}{0.9}
\definecolor{green}{HTML}{C7F6C7} 
\colorlet{green}{green!40}
\definecolor{lightblue}{HTML}{ace5ee} 
\colorlet{lightblue}{lightblue!40} 
\definecolor{lightred}{HTML}{FFC1C3} 
\colorlet{lightred}{lightred!40} 
\definecolor{lightyellow}{HTML}{FFFFAD}
\colorlet{lightyellow}{lightyellow!40} 

\definecolor{purple}{HTML}{8C6FDE}
\definecolor{blue}{HTML}{66B3FF}
\definecolor{red}{HTML}{D24B54}
\definecolor{red arrow}{HTML}{DA6C73}

\definecolor{aj_red}{HTML}{fab6bb}
\definecolor{aj_green}{HTML}{a2fada}
\definecolor{aj_blue}{HTML}{bddcff}

\definecolor{s}{HTML}{fed8ad}
\definecolor{g}{HTML}{fdfcd0}

\definecolor{medium purple}{HTML}{e6d8fc}
\definecolor{uranian blue}{HTML}{B0D5FF}

\usepackage{algorithm}
\usepackage{algpseudocode}

\usepackage[most]{tcolorbox}   	
\usepackage{float}
\usepackage{multirow}

\usepackage{booktabs}  
\usepackage{siunitx}  

\definecolor{darkblue}{rgb}{0.0,0.0,0.65}
\definecolor{darkred}{rgb}{0.65,0.0,0.0}
\definecolor{darkgreen}{rgb}{0.0,0.5,0.0}
\definecolor{tab:blue}{RGB}{31,119,180}  
\definecolor{tab:red}{RGB}{214,39,40}  
\definecolor{tab:green}{RGB}{44,160,44}  
\definecolor{tab:orange}{RGB}{255,127,14}  

\definecolor{worst}{HTML}{990000}  
\definecolor{bad}{HTML}{CC6666}    
\definecolor{neutral}{HTML}{FFCC00} 
\definecolor{good}{HTML}{4C8C4C}   
\definecolor{best}{HTML}{006600}   

\definecolor{straw}{HTML}{bab856}

\definecolor{old}{HTML}{00B050}   
\definecolor{new}{HTML}{FFC000} 

\definecolor{uie}{HTML}{1FC186} 
\definecolor{user}{HTML}{6137D2} 
 
\usepackage{subcaption}
\usepackage{array} 

\hypersetup{
    colorlinks = true,
    citecolor  = darkblue,
    linkcolor  = darkred,
    filecolor  = darkblue,
    urlcolor   = darkblue,
}
\usepackage{multirow}

\title{\textsc{C$^2$}: Scalable Auto-Feedback for LLM-based Chart Generation}

\author{
 \textbf{Woosung Koh\textsuperscript{1}}\thanks{Equal contribution, co-first authors}\thanks{Work done while an intern at KAIST AI},
 \textbf{Jang Han Yoon\textsuperscript{1}}\footnotemark[1],
  \textbf{MinHyung Lee\textsuperscript{1}},
 \textbf{Youngjin Song\textsuperscript{1}},
\\
 \textbf{Jaegwan Cho\textsuperscript{1}},
 \textbf{Jaehyun Kang\textsuperscript{1}},
 \textbf{Taehyeon Kim\textsuperscript{2}}\thanks{Mentor},
 \textbf{Se-Young Yun\textsuperscript{2}}\footnotemark[4],
\\
 \textbf{Youngjae Yu\textsuperscript{1}}\footnotemark[4],
 \textbf{Bongshin Lee\textsuperscript{1}}\thanks{Co-corresponding authors}
\\
\\
 \textsuperscript{1}Yonsei University, \textsuperscript{2}KAIST AI
\\
  \texttt{\{reiss.koh, jeffrobot99, yjy, b.lee\}@yonsei.ac.kr}
  \\
    \texttt{\{potter32, yunseyoung\}@kaist.ac.kr}
    \\
}

\begin{document}
\maketitle
\begin{abstract} 
Generating high-quality charts with Large Language Models (LLMs) presents significant challenges due to limited data and the high cost of scaling through human curation. 
$\langle \text{instruction}, \text{data}, \text{code} \rangle$ triplets are scarce and expensive to manually curate as their creation demands technical expertise.
To address this scalability challenge, we introduce a \textit{reference-free} automatic feedback generator, which eliminates the need for costly human intervention. Our novel framework, \textbf{\textsc{C$^2$}}, consists of (1) an automatic feedback provider (\textsc{\textbf{C}hartAF}) and (2) a diverse, reference-free dataset (\textsc{\textbf{C}hartUIE-8K}). The results are compelling: in our first experiment, 74\% of respondents strongly preferred, and 10\% preferred, the results after feedback. The second post-feedback experiment demonstrates that \textsc{ChartAF} outperform nine baselines. Moreover, \textsc{ChartUIE-8K} significantly improves data diversity by increasing queries, datasets, and chart types by 5982\%, 1936\%, and 91\%, respectively, over benchmarks. Finally, a  study of LLM users revealed that 94\% of participants preferred \textsc{ChartUIE-8K}'s queries, with 93\% deeming them aligned with real-world use cases. Core contributions are available as open-source at \href{http://chartsquared.github.io}{chartsquared.github.io}, with ample qualitative examples.
\end{abstract}

\section{Introduction}\label{sec:intro}

Charts are a powerful means to convey information in diverse fields, including journalism, business, and scientific research \citep{fox2011changing, rodriguez2015telling, 9012031}. 
With the success of foundation models \citep{kaplan2020scaling,roziere2023code}, there has been an increasing demand for generating charts using Large Language Models (LLMs). For instance, LIDA \citep{dibia2023lida} uses LLMs to automatically generate visualizations and infographics, and Chat2VIS \citep{10121440} incorporates LLMs to create charts from natural language queries. Moreover, LLM-generated charts empower humans by helping non-experts generate high-quality charts \citep{10121440} and improving accesibility to those with special needs \citep{gorniak2023vizability,moured2024chartformer}. Despite the rising interest, two key challenges persist: \textbf{(i)} the difficulty in evaluating LLM-generated charts 
and \textbf{(ii)} the limited availability of training data.

\textbf{(i)} Chart generation lacks straightforward evaluation methods, making it difficult to assess and improve the quality of LLM-generated charts. 
Unlike tasks with clear-cut answers, such as mathematical problem-solving where verifiers can automatically assess correctness \citep{uesato2022solving, wang2024litesearch}, chart evaluation is inherently subjective. Multiple correct designs may exist for a task (or goal), and quality often aligns with human aesthetic and functional preferences. Consequently, current evaluation systems \citep{yang2024matplotagent, wu2024plot2code, xia2024chartx} rely on reference-based approaches, necessitating labor-intensive $\langle \text{instruction}, \text{data}, \text{code}\footnote{Code here can be replaced with the image generated by executing the code.} \rangle$ triplets as a gold reference for evaluation and thus limiting their scalability. 


\begin{figure*}
    \centering
\includegraphics[width=1\linewidth]{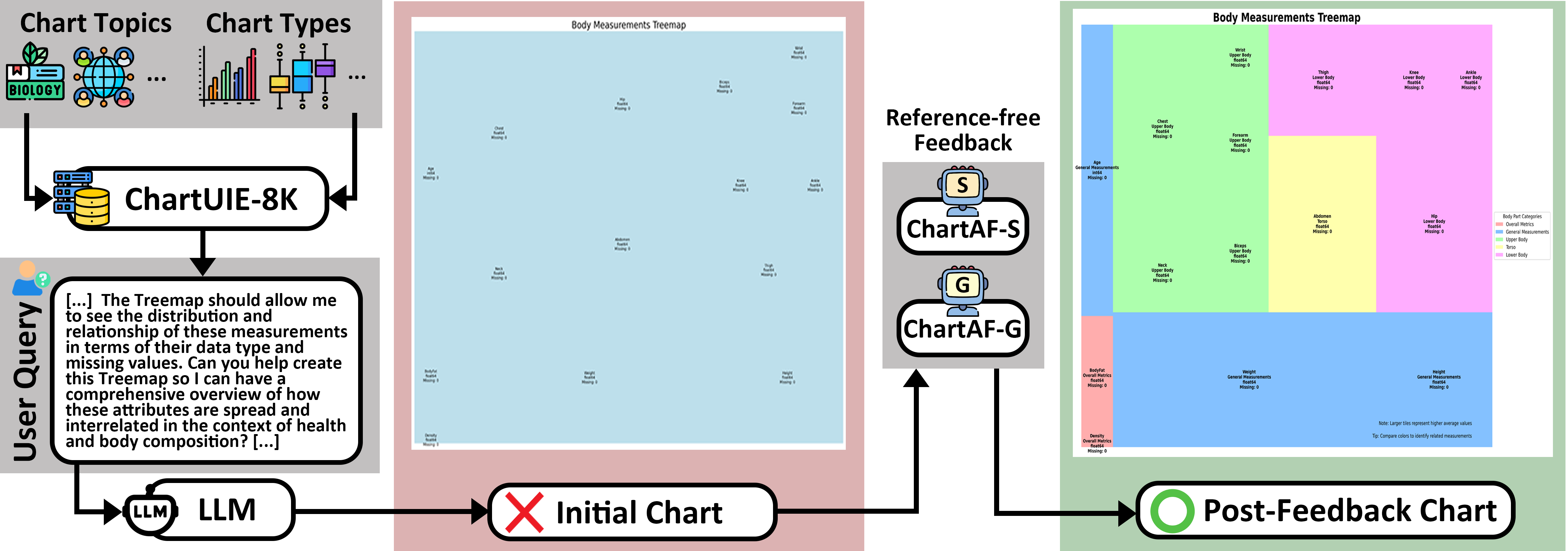} \caption{Schematic overview of \textbf{\textsc{C$^2$}} illustrating the synergy between \textsc{ChartUIE-8K} and \raisebox{0pt}{\textsc{ChartAF}}. The scale is made possible by \textsc{ChartAF}'s capability to provide \textit{reference-free} feedback. An end-to-end example is available on our \href{https://chartsquared.github.io/}{project site}'s \href{https://github.com/chartsquared/C-2}{github README}.}
\label{fig:overview}
\end{figure*}

\textbf{(ii)} Furthermore, in contrast to image generation, which typically requires only $\langle \text{instruction}, \text{image} \rangle$ pairs \citep{radford2021learning}, chart generation demands more complex $\langle \text{instruction}, \text{data}, \text{code} \rangle$ triplets. This significantly increases the costs associated with data collection and annotation. The limited number and diversity of available data restrict the variety of charts users can generate, making chart generation expensive and labor-intensive even for common applications \citep{vazquez2024llms}.



To effectively tackle both \textbf{(i)} and \textbf{(ii)}, it is essential to first address the primary bottleneck---reference-based evaluation---which also opens the door to significantly improving data diversity and scale. To this end, we introduce
\textbf{\textsc{C$^2$}}, a scalable framework, composed of the following two synergistic components:
\begin{itemize}
    \item \textsc{\textbf{C}hartAF}: A pipeline for Chart Auto-Feedback, comprising \textsc{ChartAF-S} for evaluation scores and \textsc{ChartAF-G} for granular feedback in natural language.
     \item \textsc{\textbf{C}hartUIE-8K}: A large-scale (over 8,000 instances) Chart User Interaction Emulation dataset.
\end{itemize}

Fig. \ref{fig:overview} illustrates a schematic overview of \textbf{\textsc{C$^2$}}. \textsc{ChartAF} empowers automatic (i.e., human-annotation-free) chart generation improvements (\textbf{Sec. \ref{sec:chartaf}}). \textsc{ChartAF} works exceptionally well reference-free, enabling the cost-effective curation of a large-scale chart generation evaluation set \textsc{ChartUIE-8K} (\textbf{Sec. \ref{sec:chartuie}}).

The quantitative and qualitative results of \textbf{\textsc{C$^2$}} is overwhelmingly positive. First, by leveraging \textsc{ChartAF}'s scalar evaluation scores for a simple test-time scaling scheme, 84\% of respondents preferred the post-feedback results, with 74\% strongly preferring them and 10\% preferring them (\textbf{Sec. \ref{sec:testscale}}). Second, employing \textsc{ChartAF}'s granular feedback to in-context tune, \textsc{ChartAF}'s post-feedback preference scores beat 9 baselines, that are alternatives to \textsc{ChartAF} (\textbf{Sec. \ref{sec:context}}). The qualitative improvements for both test-time scaling and in-context tuning can be viewed on our open-source \href{http://chartsquared.github.io}{project site}. 

Finally, \textsc{ChartAF}'s reference-free nature allows us to curate, \textsc{ChartUIE-8K}, dramatically raising data diversity via number of queries, underlying datasets, and chart types by 5982\%, 1936\%, and 91\%, respectively, against existing evaluation sets (\textbf{Tab. \ref{tab:uie}}). We also demonstrate that \textsc{ChartUIE-8K} closely aligns with real-world human requests via a study of LLM users (\textbf{Sec. \ref{sec:uie_exp}}). The study highlights that \textsc{ChartUIE-8K}'s evaluation set distribution closely aligns with real-world users, and 94\% and 93\%, prefer, and think is realistic, respectively.

\subsection{Related Work}\label{sec:related}
\paragraph{\textbf{Chart Generation.}} \citet{10121440} offer prompt-engineered, LLM-based chart generation---however, they only provide qualitative case studies to verify their contribution. \citet{dibia2023lida} presents an LLM-based infographic visualization tool, which includes interactive charts. However, \citet{dibia2023lida} does not include a human study, making it challenging to assess its effectiveness. \citet{sah2024generatinganalyticspecificationsdata} propose a natural language-to-chart recommendation approach based on the visualization language Vega-Lite. Therefore, their task deviates from the LLM-based chart generation we tackle. \citet{Tian_2024} recently proposes a chart generation work that generates charts from "abstract user utterances" (as stated in the paper), which diverges from the instruction-based queries we address. An example of an "abstract user utterance" they provide is "What kind of movies earn the most recently?"—this diverges from the example instructions provided in our study of LLM users (App. \ref{sec:user_inst}). This difference is understandable as their chart generation is based on proprietary user-interaction software, not a common LLM chatbot.

\paragraph{Tuning with Feedback.} Feedback-based LLM tuning is commonly done via three methods: parameter tuning \citep{ouyang2022training}, in-context tuning \citep{liu2022p}, and test-time scaling (TTS; \citet{openaio1}). Parameter tuning occurs through (implicit or explicit) rewards, requiring large amounts of reward data. For instance, \citet{kirstain2023pick} and \citet{xu2024imagereward} collected over 500,000 and 137,000 annotations, respectively. In-context tuning allows LLMs to improve via $n$-shot generation \citep{NEURIPS2023_398b00a0}, which requires $n$ additional prompts. TTS refers to the case where a verifier with an ordinal output (typically of scalar value) can help improve the final generation \citep{snell2024scaling}. Each tuning method requires an external feedback-provider, such as a reward model, additional prompt, or verifier.

\section{\textbf{\textsc{C$^2$}}: \textsc{ChartAF}} \label{sec:chartaf}
In \textbf{\textsc{C$^2$}}, a feedback-provider, \textsc{ChartAF} enables TTS and in-context tuning. 
We first describe the shortcomings of existing approaches, and introduce  two versions of \textsc{ChartAF}: \textsc{ChartAF-S} and \textsc{ChartAF-G}.

\subsection{Towards High-performing Feedback}\label{sec:towards}

Two feedback types have been explored in the LLM research community: \textbf{(1)} scalar score-based ($s_\mathbb{N} \in \mathbb{N}_{\cup 0}$) and \textbf{(2)} natural language-based ($s_\mathcal{N} \in \mathcal{N}$, where $\mathcal{N}$ is the natural language set space) feedback. Instead of competing, \textbf{(1)} and \textbf{(2)} serve different purposes: \textbf{(1)} is suited for parameter tuning with rewards, and  TTS, while \textbf{(2)} is suited for in-context tuning. We summarize prior efforts of the two approaches in chart generation below.

\begin{figure*}
    \centering
\includegraphics[width=1\linewidth]{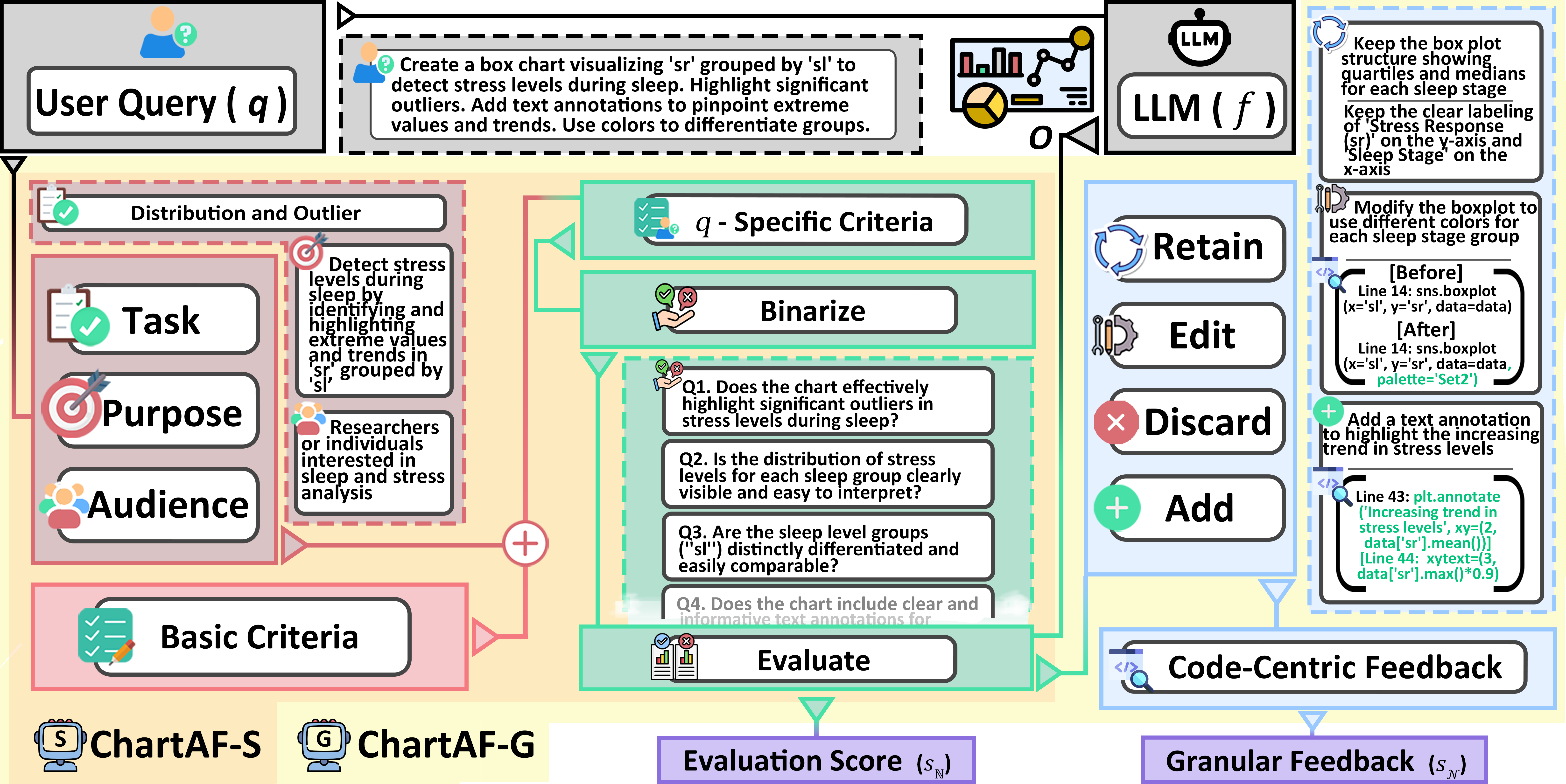}
\caption{Schematic diagram of \textsc{ChartAF}, including a qualitative example indicated by dashed containers. The process starts in the top-left with the user query, which is processed by the chart-generating LLM and either \raisebox{0pt}{\colorbox{s}{\textsc{ChartAF-S}}} or \raisebox{0pt}{\colorbox{g}{\textsc{ChartAF-G}}}. Notably, \raisebox{0pt}{\colorbox{s}{\textsc{ChartAF-S}}} and \raisebox{0pt}{\colorbox{g}{\textsc{ChartAF-G}}} both share the first two modules (\raisebox{0pt}{\colorbox{aj_red}{red}} and \raisebox{0pt}{\colorbox{aj_green}{green}}). The final output is a scalar evaluation score or granular feedback, depending on the chosen path.}
\label{fig:chartaf}
\end{figure*}
\textbf{(1)} \citet{yang2024matplotagent}, \citet{wu2024plot2code}, and \citet{xia2024chartx} provide $[0,100]$, $[1,10]$, $[0,5]$, scores, respectively. However, as these methods were developed under a reference-based regime, their performance sharply deprecates when applied reference-free. While there are other similar works, they are closed-source without clear details for replication \citep{han2023chartllama,xu2023chartbench}. 

\textbf{(2)} A naïve approach uses a zero-shot LLM to provide feedback \citep{yang2024matplotagent}, which we call Naïve Feedback (NF). This method, as our experiments later show, is ineffective (Tab. \ref{table:exp1}). Furthermore, \citet{yang2024matplotagent} does not provide human studies of their NF against baselines so there is no evidence on its efficacy. 

To overcome the limitations of past works, we present \textsc{ChartAF} (Fig. \ref{fig:chartaf}). As \textsc{ChartAF-S} is a subset of \textsc{ChartAF-G}, we first introduce \textsc{ChartAF-S} and then the additional component corresponding to \textsc{ChartAF-G}. Fine-grained pseudocode and prompts are provided in App. \ref{app:chartaj}.

\subsection{\raisebox{0pt}{{\textsc{ChartAF-S} ($\tilde{f}_{AF}$)}}} \label{sec:s}
\paragraph{\raisebox{0pt}{{\textbf{Module 1.}}}} The user query, $q \in \mathcal{Q} \in \mathcal{N}$, is first decomposed into three essential factors in a chart generation query: Task, Purpose, and Audience 
\citep{10.1145/3154862.3154881,lee2020reaching,narechania2020nl4dv,wang2020cheat,parsons2021understanding}. By explicitly decomposing the $q$ into Task, Purpose, and Audience, \textsc{ChartAF} can better infer the user intention \citep{quadri2024you,bressa2024input}. Since user queries are often brief (Fig. \ref{fig:uiestudy}, App. \ref{sec:user_inst}), the intention may not always be clearly stated. Nevertheless, \textsc{ChartAF} utilizes the $q$ to induce underlying intentions. 

This information is then fused with the Basic Criteria---a general, high-level criteria applicable to all chart evaluations. The Basic Criteria ensures that \textsc{ChartAF} comprehensively considers chart elements: Chart Type, Visual Embellishment, Text, Color, Annotation, Aesthetics, and Visual Clutter. This criteria is inspired by the rich literature in visualization research. We document the research that corresponds to each element in App. \ref{app:ref}.

This first module is the domain grounding module. \textsc{ChartAF} replaces gathering costly human annotations with existing scholarly research. Not only is this cheaper, this approach closely aligns with  how human-made chart generations would be evaluated. We would grade human students' chart generations with a domain expert (lecturer) that has learned the principles of chart generation, grounded in scholarly literature \citep{bach2023challenges}.

\paragraph{\raisebox{0pt}{{\textbf{Module 2.}}}} Considering the decomposed $q$, and Basic Criteria, \textsc{ChartAF} generates $q$-Specific Criteria---specializing the general Basic Criteria. Specialization is key to generating feedback that is customized to the specific $q$ \citep{10.1145/3613904.3642216}. The criteria is then transformed to binary (i.e., yes or no) questions \citep{hu2023tifa} as we find that LLMs are more reliable when reasoning with binary rather than open-ended questions. This also allows \textsc{ChartAF} to explicitly associate each criterion with one question. Otherwise, LLMs tend to lump numerous criterion together, resulting in duplicate related criteria. 

Then, these questions are evaluated considering the LLM-generated chart (generated by executing the code output $o \in \mathcal{O} \in \mathcal{N}$). This code is generated by the chart generating LLM, $f: \mathcal{Q} \mapsto \mathcal{O}$. The final scalar, $s_\mathbb{N}$, is derived by equal-weighting each binary answer, yes (1) and no (0). If researchers only require a scalar score, for downstream applications, the process can be terminated.

\subsection{\raisebox{0pt}{{\textsc{ChartAF-G} ($f_{AF}$)}}}\label{sec:g}
\paragraph{\raisebox{0pt}{{\textbf{Module 3.}}}} For granular feedback, the process continues by associating each answered question with Retain, Edit, Discard, and Add. This association helps to decompose the evaluation result of each criterion to \textit{actionable} feedback. Based on this association, code-centric text feedback is provided \citep{bi2024iterativerefinementprojectlevelcode}: the LLM is prompted to provide fine-grained feedback considering the downstream application, i.e., chart generation via code generation. Code-centric feedback encourages the LLM to provide explicit feedback that can be directly applied to the downstream $f$.

\section{\textbf{\textsc{C$^2$}}: \textsc{ChartUIE-8K}}\label{sec:chartuie}
Leveraging the reference-free nature of \textsc{ChartAF}, we curate \textsc{ChartUIE-8K}, a comprehensive chart generation evaluation set. Since no gold references are required, we can significantly scale the dataset. Qualitative examples of \textsc{ChartUIE-8K} can be viewed in Figs. \ref{fig:overview}, \ref{fig:chartaf}, App. \ref{app:query}, and our \href{https://chartsquared.github.io/}{project site}.

\begin{figure}
    \centering
\includegraphics[width=1\linewidth]{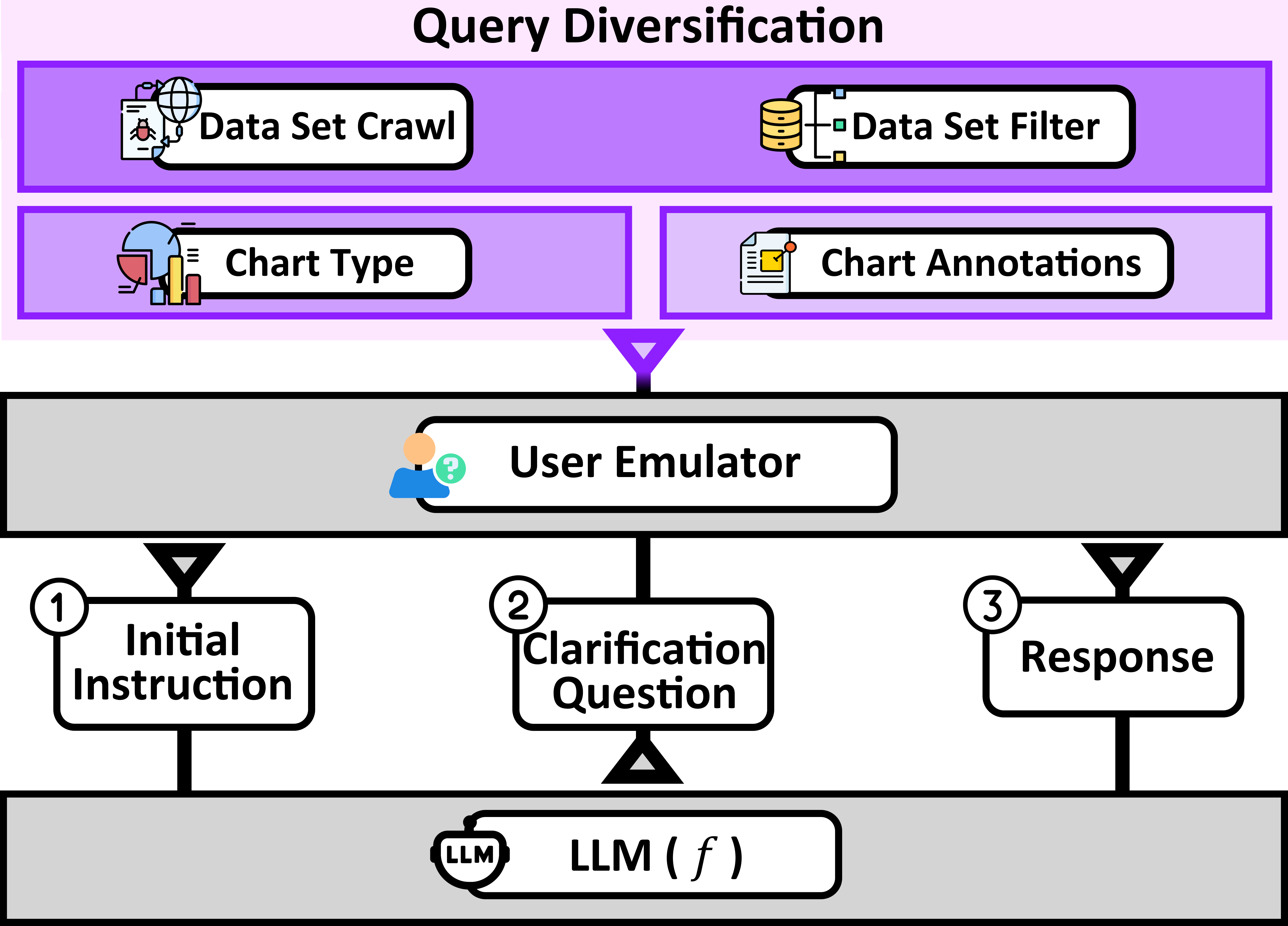}
    \caption{\textsc{ChartUIE-8K} curation schematic diagram.}
    \label{fig:uie}
\end{figure}

\paragraph{\textbf{Curation Method.}} An overview of \textsc{ChartUIE-8K}'s curation process is illustrated in Fig. \ref{fig:uie}. First, for diverse chart topics, we semi-automatically crawl diverse datasets online with appropriate licenses. We include only the datasets that have been used by humans for chart generation purposes, as not all datasets are suitable for visualization. Next, to ensure diverse chart types and annotations, we adopt a comprehensive list of chart types \citep{2022types} and annotations \citep{ren2017chartaccent}. Then, we emulate two types of users: [\textbf{U1}] lay users and [\textbf{U2}] detailed users.

\begin{figure*}
    \centering
    \begin{subfigure}[b]{0.495\textwidth}
        \centering
        \includegraphics[width=\textwidth]{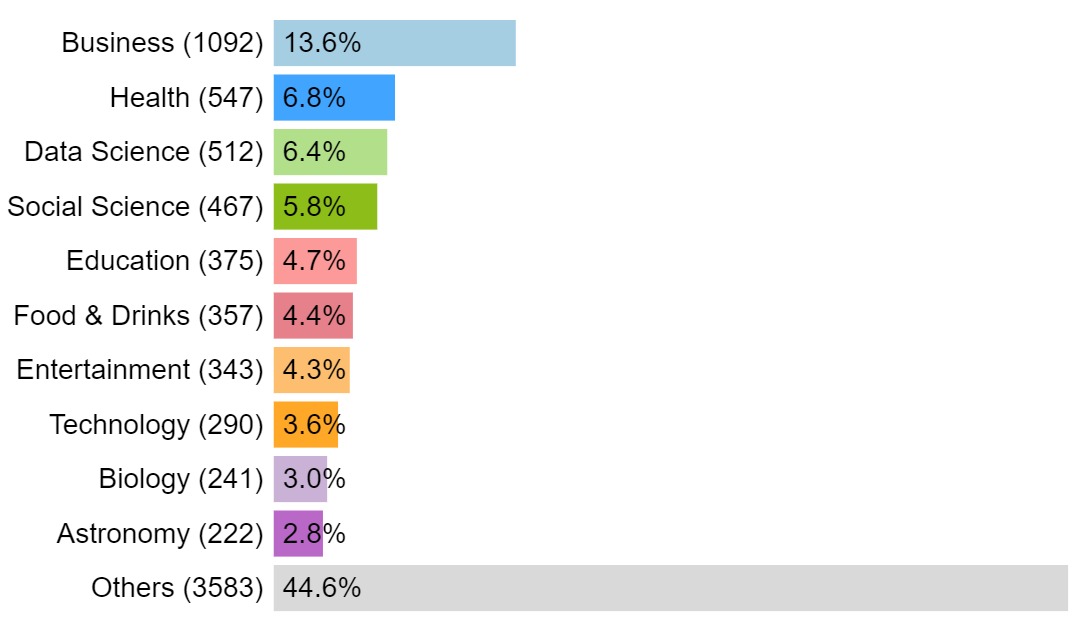}
        \caption{Chart topics in order of frequency}
        \label{fig:topic}
    \end{subfigure}
    \begin{subfigure}[b]{0.495\textwidth}
        \centering
        \includegraphics[width=\textwidth]{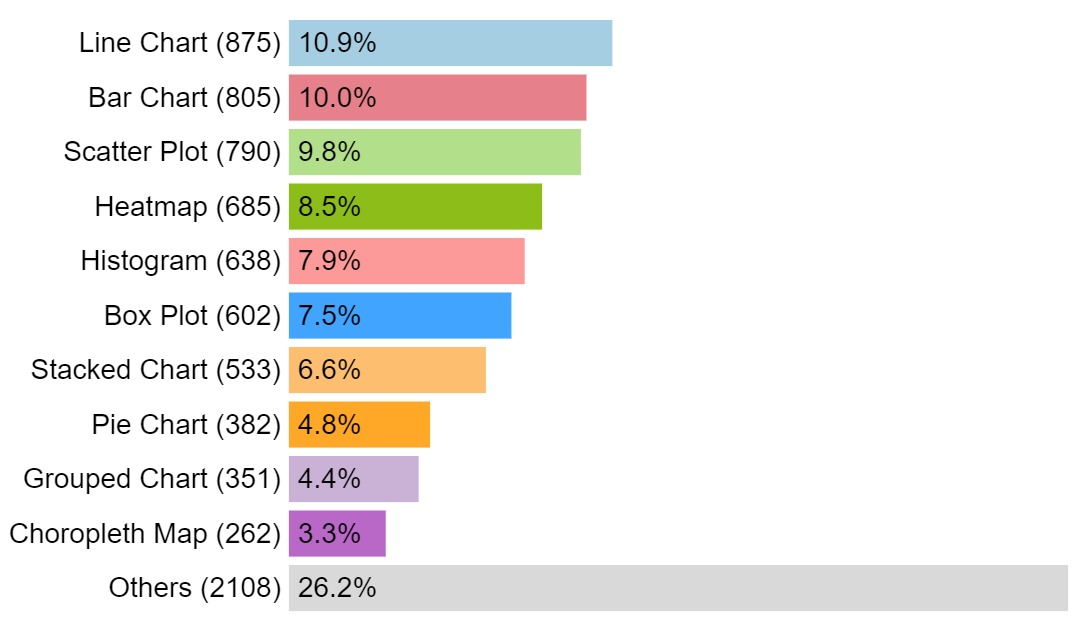}
        \caption{Chart types in order of frequency}
        \label{fig:type}
    \end{subfigure}
\caption{\textsc{ChartUIE-8K} distribution. Top 10 topics and types are explicitly depicted while the remaining is classified as others. ($n$) is the number of samples out of 8028.}
    \label{fig:dist}
\end{figure*}

To this end, we synthetically (LLM-assisted) generate initial instructions with two configurations: approx. [\textbf{U1}] <50, and [\textbf{U2}] <100 words. Considering the salience of multi-turn benchmarking \citep{wang2024mint}, we further emulate a single QA cycle. I.e., the LLM asks for clarifying questions to the user, then, the user emulator responds to [\textbf{U1}] 25\% or [\textbf{U2}] 50\% of the questions. The pseudocode is provided in App. \ref{app:uie_pseudo}.

\paragraph{\textbf{Statistical Summary.}} To systematically understand the evaluation set, we compare key statistics against relevant benchmarks (see Tab. \ref{tab:uie}). We exclude chart topic counts for benchmarks, as they are not provided in MatPlotBench \citep{yang2024matplotagent} and Plot2Code \citep{wu2024plot2code}. We do not manually count their chart topics as this a subjective task. On the other hand, we count MatPlotBench's and Plot2Code's chart types using the same taxonomy used for \textsc{ChartUIE-8K} (App. \ref{app:uie_pseudo}). We leave the original number of coarse chart types provided by Plot2Code in parentheses for documentation. Finally, the distribution of chart topics and types of \textsc{ChartUIE-8K} are presented in Fig. \ref{fig:dist}. 

\begin{table}
    \centering
  \resizebox{\columnwidth}{!}{
    \begin{tabular}{llll}
    \toprule
        Number of & \textsc{\textbf{ChartUIE-8K}} & \textbf{MatPlotBench} & \textbf{Plot2Code}\\
    \midrule
        Queries & 8028 (\textbf{+5982\%})& 100 & 132 \\
        Datasets & 509 (\textbf{+1936\%})& 25& 2\\
        Chart topics & 44& --- & --- \\
        Chart types & 63 (\textbf{+91\%})& 33& 24 (6)\\
    \bottomrule
    \end{tabular}}
    \caption{Comparing key statistics of chart generation evaluation sets. \textbf{Bold} represents improvement from the best existing benchmark.}
    \label{tab:uie}
\end{table}

\section{Empirical Study}\label{sec:exp}
\subsection{Preliminary and Notations}\label{sec:prelim}

Denote the chart code-generating LLM, $f: \mathcal{Q} \mapsto \mathcal{O}$. Let a feedback-provider take the output of $f$, including the executed chart image, then $\tilde{f}_{AF}: \mathcal{O} \mapsto \mathcal{S}_\mathbb{N} \in \mathbb{N}_{\cup 0}$, $f_{AF}: \mathcal{O} \mapsto \mathcal{S}_\mathcal{N} \in \mathcal{N}$, for TTS, and in-context tuning feedback, respectively. Following \citet{liang2024rich}, $h: \mathcal{O}_{pre} \times \mathcal{O}_{post} \mapsto \{-2, -1, 0, 1, 2\}$ represents the human post-feedback preference score function. The co-domain refers to $\succ$: strongly prefer pre-feedback (-2), $\succeq$: prefer pre-feedback (-1), $\sim$: indifferent (0), $\preceq$: prefer post-feedback (1), $\prec$: strongly prefer post-feedback (2). To demonstrate \textsc{ChartAF-S} ($\tilde{f}_{AF}$) and \textsc{ChartAF-G} ($f_{AF}$)'s utility, we empirically study TTS with \textsc{ChartAF-S} (\textbf{Sec. \ref{sec:testscale}}) and in-context tuning with \textsc{ChartAF-G} (\textbf{Sec. \ref{sec:context}}). Fine-grained details are reported in App. \ref{app:humanstudy}.
\begin{figure}
    \centering
\includegraphics[width=1\linewidth]{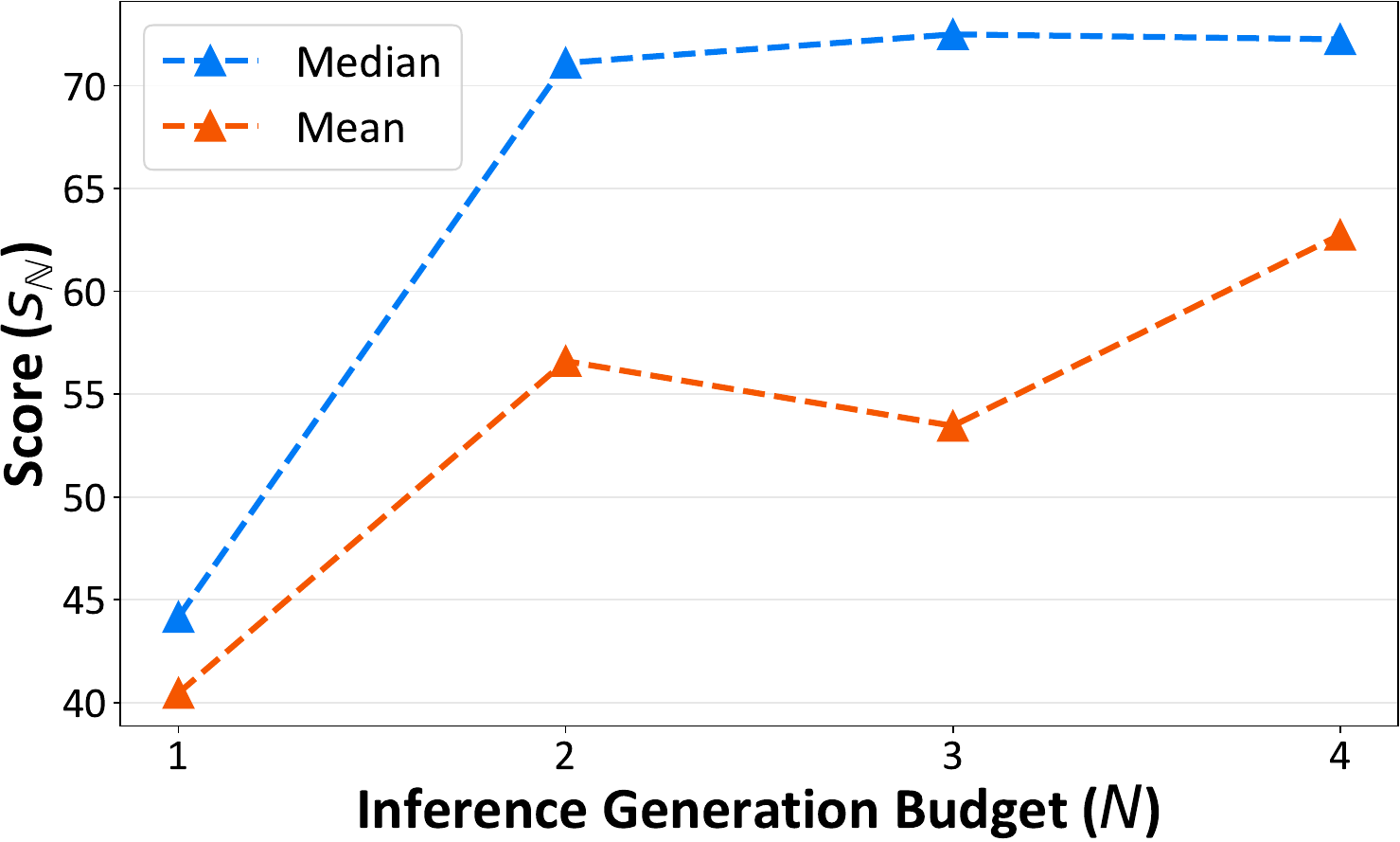}
    \caption{Test-time scaling with \textsc{ChartAF} as a verifier. Raising $N$ leads to improved generations.}
    \label{fig:tts}
\end{figure}

\subsection{Test-time Scaling with \textsc{ChartAF}}\label{sec:testscale}
\paragraph{\textbf{Experiment Set-up.}} We demonstrate that \textsc{ChartAF-S} is an effective verifier for TTS. Following \citet{snell2024scaling}'s parallel best-of-$N$, we generate $N \in \mathbb{N}$ independent samples and choose the one with the highest score ($s_\mathbb{N} \in \mathcal{S}_\mathbb{N} := [0,100]$). Here, the unit of inference budget is $N$s. Let memory set $\mathcal{M}:=\{o_1, \cdots, o_N\}$, $o\in \mathcal{O}$, hold the $N$ independent outputs of $f$. Then, we experimentally show that 
\begin{equation}
\hat{o} := \underset{o \ \in \ \mathcal{M}}{\text{arg max}} \ \tilde{f}_{AF}(o),
\label{eq:tts1}\end{equation}
\begin{equation*}
\text{if} \ \hat{o}_i > \hat{o}_j, \ i \neq j \in \mathbb{N},
\end{equation*}
\begin{equation} \label{eq:tts}
    \Rightarrow h(\hat{o}_{j:=1}:=f(q), \hat{o}_i) > 0,
\end{equation}
across $q \in \mathcal{Q}$.

We employ a double-blind 120 human study (of which 77 pass the rigorous sanity check) to compare pre-TTS ($N := 1$) and post-TTS ($N := 4$) preference scores. Each participant is presented with a random TTS sample. For this experiment we set both $f$ and $\tilde{f}_{AF}$ backbone as \texttt{GPT 4o} to emphasize that $\tilde{f}_{AF}$ does not have to be a superior model for \textsc{ChartAF-S} to be useful.

\paragraph{\textbf{Results.}} First, the scaling curve is depicted in Fig. \ref{fig:tts}. The positive slope highlights that \textsc{ChartAF} can be effectively used as a TTS verifier. To rigorously verify this claim, the distribution of the human study is presented in Tab. \ref{table:tts}. Going from a median $s_\mathbb{N}$ of 40.47$\rightarrow$62.75 (pre-$\rightarrow$post-TTS) leads to 74\% strongly preferring the post-TTS and 10.4\% preferring post-TTS. This is a strong indication that $s_\mathbb{N}$ is closely proxying human preferences.

\begin{table}
    \centering
  \resizebox{\linewidth}{!}{
    \begin{tabular}{ccccc}
    \toprule
       \cellcolor{worst!20} $\succ$ & \cellcolor{bad!20}$\succeq$ & \cellcolor{neutral!20}$\sim$ & \cellcolor{good!20}$\preceq$ & \cellcolor{best!20}$\prec$ \\
    \midrule
\cellcolor{worst!20} 3 (\textbf{3.9\%}) & \cellcolor{bad!20} 4 (\textbf{5.2\%}) & \cellcolor{neutral!20} 5 (\textbf{6.5\%}) & \cellcolor{good!20} 8 (\textbf{10.4\%}) & \cellcolor{best!20} 57 (\textbf{74\%}) \\
    \bottomrule
    \end{tabular}
    }
    \caption{Number of respondents out of 77 (\textbf{\%}). $\succ$: strongly prefer pre-TTS, $\succeq$: prefer pre-TTS, $\sim$: indifferent, $\preceq$: prefer post-TTS, $\prec$: strongly prefer post-TTS.}
    \label{table:tts}
\end{table}

\subsection{In-context Tuning with \textsc{ChartAF}}\label{sec:context}
\paragraph{\textbf{Experiment Set-up.}} For a comprehensive empirical study we consider four LLMs: \textbf{(i)} \texttt{GPT 4o} \citep{achiam2023gpt}, \textbf{(ii)} \texttt{Claude 3.5 Sonnet} \citep{claude3}, \textbf{(iii)} \texttt{Llama 3.1 70b} \citep{dubey2024llama}, and \textbf{(iv)} \texttt{Gemma 2 27B} \citep{team2024gemma}---of which \textbf{(i)} and \textbf{(ii)} are closed-source and \textbf{(iii)} and \textbf{(iv)} are open-source. The two closed-source models are used as $f_{AF}$ backbones, while all four models are used as $f$. Detailed LLM configurations are provided in App. \ref{app:config}.
\begin{table*}
\centering
  \caption{Empirical study preference scores as presented in Sec. \ref{sec:prelim}. $\mu$ represents mean, $\tilde{\mu}$ represents median. \textbf{Bold} is the best result across the column, and \underline{underlined} is the second-best. Total (input plus output) token costs per query evaluation are reported as $\mu \pm \sigma$.}
  \resizebox{0.95\textwidth}{!}{
  \begin{tabular}{|l|rr|rr|rr|rr|c|}
    \toprule
        $f$ & \multicolumn{2}{c|}{\texttt{GPT 4o}} & \multicolumn{2}{c|}{\texttt{Claude 3.5 Sonnet}} & \multicolumn{2}{c|}{\texttt{Llama 3.1 70b}} & \multicolumn{2}{c|}{\texttt{Gemma 2 27B}} &$f_{AF}$\\
        \cmidrule{1-9}
        Statistic &$\mu$&$\tilde{\mu}$&$\mu$&$\tilde{\mu}$&$\mu$&$\tilde{\mu}$&$\mu$&$\tilde{\mu}$&Token Cost\\
         \midrule
         $f_{AF}$ Backbone &\multicolumn{9}{c|}{\texttt{GPT 4o}} \\
         \midrule
   \cellcolor{gray!20}\textsc{ChartAF} (\textbf{ours}) & \cellcolor{good!20}\underline{0.5} & \cellcolor{good!20}\textbf{1} &\cellcolor{good!20}\textbf{0.5} & \cellcolor{best!20}\textbf{1.5} & \cellcolor{good!20}\textbf{1.308} & \cellcolor{best!20}\textbf{2} & \cellcolor{good!20}\textbf{0.333} & \cellcolor{good!20}\textbf{1} & 14013$_{\pm 1155}$ \\
      ChatEval & \cellcolor{bad!20}-0.375 & \cellcolor{bad!20}-1 & \cellcolor{bad!20}-0.583 & \cellcolor{bad!20}-0.5 & \cellcolor{bad!20}-0.615 & \cellcolor{bad!20}-1 & \cellcolor{bad!20}-1.222 & \cellcolor{worst!20}-2 & 70195$_{\pm 8121}$ \\
   ChartX+A-CoT+SC & \cellcolor{bad!20}-0.25 & \cellcolor{bad!20}-1 & \cellcolor{bad!20}-0.167 & \cellcolor{neutral!20}0 & \cellcolor{bad!20}-1.154 & \cellcolor{worst!20}-2 & \cellcolor{good!20}\underline{0.111} & \cellcolor{good!20}\textbf{1} & 31791$_{\pm 4604}$ \\
   MatPlotBench+A-CoT+SC & \cellcolor{bad!20}-0.25 & \cellcolor{neutral!20}0 & \cellcolor{bad!20}-0.5 & \cellcolor{bad!20}-1 & \cellcolor{bad!20}-0.231 & \cellcolor{neutral!20}\underline{0} & \cellcolor{bad!20}-0.556 & \cellcolor{bad!20}-1 & 24773$_{\pm 2149}$ \\
   Plot2Code+A-CoT+SC & \cellcolor{bad!20}-0.75 & \cellcolor{worst!20}-1.5 & \cellcolor{bad!20}-0.25 & \cellcolor{neutral!20}0 & \cellcolor{bad!20}-0.385 & \cellcolor{neutral!20}\underline{0} & \cellcolor{bad!20}-1 & \cellcolor{bad!20}-1 & 22519$_{\pm 2122}$ \\
   ChartX  & \cellcolor{good!20}\underline{0.5} & \cellcolor{good!20}\underline{0.5} & \cellcolor{good!20}\underline{0.083} & \cellcolor{good!20}\underline{1} & \cellcolor{bad!20}-0.385 & \cellcolor{neutral!20}\underline{0} & \cellcolor{bad!20}-1.111 & \cellcolor{worst!20}-2 & 1665$_{\pm 235}$ \\
   MatPlotBench & \cellcolor{good!20}0.25 & \cellcolor{good!20}\underline{0.5} & \cellcolor{bad!20}-0.5 & \cellcolor{bad!20}-1 & \cellcolor{bad!20}-0.385 & \cellcolor{neutral!20}\underline{0} & \cellcolor{bad!20}-0.444 & \cellcolor{neutral!20}\underline{0} & 1653$_{\pm 284}$ \\
   Plot2Code & \cellcolor{bad!20}-0.5 & \cellcolor{bad!20}-0.5 & \cellcolor{bad!20}-0.167 & \cellcolor{bad!20}-0.5 & \cellcolor{bad!20}\underline{-0.077} & \cellcolor{neutral!20}\underline{0} & \cellcolor{bad!20}-0.667 & \cellcolor{bad!20}-1 & 1604$_{\pm 277}$ \\
   NF & \cellcolor{good!20}\textbf{0.75} & \cellcolor{good!20}\textbf{1} & \cellcolor{bad!20}-0.583 & \cellcolor{bad!20}-1 & \cellcolor{bad!20}-0.692 & \cellcolor{worst!20}-2 & \cellcolor{bad!20}-0.333 & \cellcolor{worst!20}-2 & 2739$_{\pm 385}$ \\
   Skip+A-CoT & \cellcolor{bad!20}-1.375 & \cellcolor{worst!20}-2 & \cellcolor{bad!20}-1 & \cellcolor{bad!20}-1 & \cellcolor{worst!20}-1.692 & \cellcolor{worst!20}-2 & \cellcolor{worst!20}-1.667 & \cellcolor{worst!20}-2 &---\\
   \toprule
     $f_{AF}$ Backbone &\multicolumn{9}{c|}{\texttt{Claude 3.5 Sonnet}} \\
    \midrule
   \cellcolor{gray!20}\textsc{ChartAF} (\textbf{ours}) & \cellcolor{good!20}\textbf{0.167} & \cellcolor{neutral!20}\underline{0} & \cellcolor{good!20}\underline{0.273} & \cellcolor{good!20}\textbf{1} & \cellcolor{good!20}\textbf{1.273}
 & \cellcolor{best!20}\textbf{2} & \cellcolor{good!20}\textbf{1} & \cellcolor{best!20}\textbf{2} & 14927$_{\pm 981}$ \\
    ChatEval & \cellcolor{neutral!20}\underline{0} & \cellcolor{good!20}\textbf{0.5} & \cellcolor{bad!20}-1 & \cellcolor{bad!20}-1 & \cellcolor{neutral!20}0 & \cellcolor{neutral!20}0 & \cellcolor{good!20}\underline{0.143} & \cellcolor{good!20}\underline{1} & 74369$_{\pm 8397}$ \\
   ChartX+A-CoT+SC & \cellcolor{bad!20}-0.667 & \cellcolor{bad!20}-1 & \cellcolor{bad!20}-0.727 & \cellcolor{worst!20}-2 & \cellcolor{good!20}\underline{0.818} & \cellcolor{good!20}\underline{1} & \cellcolor{bad!20}-0.571 & \cellcolor{bad!20}-1 & 34210$_{\pm 4084}$ \\
   MatPlotBench+A-CoT+SC & \cellcolor{bad!20}-1.333 & \cellcolor{worst!20}-1.5 & \cellcolor{good!20}\underline{0.273} & \cellcolor{good!20}\textbf{1} & \cellcolor{bad!20}-0.455 & \cellcolor{bad!20}-1 & \cellcolor{bad!20}-1 & \cellcolor{worst!20}-2 & 28980$_{\pm 2608}$ \\
   Plot2Code+A-CoT+SC  & \cellcolor{bad!20}-0.667 & \cellcolor{worst!20}-1.5 & \cellcolor{bad!20}-0.727 & \cellcolor{bad!20}-1 & \cellcolor{good!20}0.273 & \cellcolor{good!20}\underline{1} & \cellcolor{bad!20}-0.286 & \cellcolor{neutral!20}0 & 26849$_{\pm 2520}$ \\
   ChartX  & \cellcolor{bad!20}-0.167 & \cellcolor{neutral!20}\underline{0} & \cellcolor{bad!20}-0.364 & \cellcolor{neutral!20}\underline{0} & \cellcolor{bad!20}-0.545 & \cellcolor{bad!20}-1 & \cellcolor{bad!20}-1.143 & \cellcolor{bad!20}-1 & 1753$_{\pm 277}$ \\
   MatPlotBench & \cellcolor{bad!20}-0.5 & \cellcolor{bad!20}-0.5 & \cellcolor{good!20}\textbf{0.364} & \cellcolor{neutral!20}\underline{0} & \cellcolor{bad!20}-0.545 & \cellcolor{bad!20}-1 & \cellcolor{neutral!20}0 & \cellcolor{neutral!20}0 & 2040$_{\pm 407}$ \\
   Plot2Code & \cellcolor{bad!20}-1.167 & \cellcolor{worst!20}-1.5 & \cellcolor{bad!20}-0.273 & \cellcolor{neutral!20}\underline{0} & \cellcolor{good!20}0.182 & \cellcolor{neutral!20}0 & \cellcolor{bad!20}-0.143 & \cellcolor{neutral!20}0 & 1851$_{\pm 412}$ \\
   NF & \cellcolor{bad!20}-1.167 & \cellcolor{worst!20}-1.5 & \cellcolor{worst!20}-1.727 & \cellcolor{worst!20}-2 & \cellcolor{neutral!20}0 & \cellcolor{good!20}\underline{1} & \cellcolor{worst!20}-1.571 & \cellcolor{worst!20}-2 & 2739$_{\pm 562}$ \\
   Skip+A-CoT & \cellcolor{bad!20}-1 & \cellcolor{bad!20}-1 & \cellcolor{bad!20}-1.182 & \cellcolor{worst!20}-2 & \cellcolor{worst!20}-1.636 & \cellcolor{worst!20}-2 & \cellcolor{worst!20}-1.571 & \cellcolor{worst!20}-2 & ---\\
   \bottomrule
   \multicolumn{10}{l}{
      \footnotesize{
        \raggedright{
        Color-coded: \colorbox{worst!20}{$\mu, \tilde{\mu} \leq -1.5$}; 
        \colorbox{bad!20}{$-1.5 < \mu, \tilde{\mu} < 0$}; 
        \colorbox{neutral!20}{$\mu, \tilde{\mu} = 0$}; 
        \colorbox{good!20}{$0 < \mu, \tilde{\mu} < 1.5$}; 
        \colorbox{best!20}{$\mu, \tilde{\mu} \geq 1.5$}
        }
      }
   } \\
  \end{tabular}
  }
  \label{table:exp1}
\end{table*}

Our empirical study employs a double-blind human study of 60 queries with 120 participants, 77 of whom pass the rigorous sanity check. 120 individuals are required for 60 queries as we test the two $f_{AF}$ independently. Within the 60 queries, 15 are designated for each of the four $f$. For representative sampling, we random sample the 15 queries with two constraints: (1) each of the 15 queries ask for different chart types, (2) 1:1 ratio of [\textbf{U1}] and [\textbf{U2}] queries. Concretely, our experiments show 
\begin{equation*}
h(f(q), f(q \oplus f_{AF}(q))) >
\end{equation*}
\begin{equation} h(f(q), f(q \oplus f^{baseline}_{AF}(q)))\label{eq:pref}\end{equation}
across $q \in \mathcal{Q}$, where $\oplus$ denotes string concatenation.

\paragraph{\textbf{Baselines.}} We identify four baselines: \textbf{(1)} ChartX \citep{xia2024chartx}, \textbf{(2)} MatPlotBench \citep{yang2024matplotagent}, \textbf{(3)} Plot2Code \citep{wu2024plot2code}, and \textbf{(4)} ChatEval \citep{chan2024chateval} that can provide feedback for chart generation. \textbf{(1) (2) (3)} are chart generation specific feedback providers, while \textbf{(4)} is a generalist. One advantage of \textbf{(1) (2) (3)} is its token cost light nature. Therefore, for a fair comparison vis-à-vis token cost, we enhance these baselines with Auto-Chain-of-Thought (A-CoT) \citep{zhang2023automatic} and Self-Consistency (SC) \citep{wang2023selfconsistency}. Beforehand, we ran preliminary studies with A-CoT, SC, and Self-Refine \citep{madaan2023selfrefine}, and found that A-CoT+SC performed best. To solidify the utility of \textsc{ChartAF} we include two additional baselines. The first is Naïve Feedback (NF) which is zero-shot asking another LLM to give feedback \citep{yang2024matplotagent}. The second is skipping the feedback stage, and instead directly adding A-CoT to $f$ (Skip+A-CoT).

\paragraph{\textbf{Results.}}
We present our findings in Tab. \ref{table:exp1} including (input plus output) token costs. \textsc{ChartAF} ranks \textbf{first} 13 (out of 16) times and \underline{second} for the remaining three. This level of consistency across four different models accentuates \textsc{ChartAF}'s utility across large and smaller $f$. Furthermore, regardless of $f_{AF}$ backbone, post-feedback always (8 out of 8) results in improvement, $\mu > 0$, spotlighting \textsc{ChartAF}'s universality. Detailed qualitative examples are provided on the \texttt{\href{chartsquared.github.io}{project site}}.

It is important to note that \textsc{ChartAF}'s distinguished performance cannot be trivially matched by enhancing baselines with state-of-the-art prompting methods. While the three enhanced baselines use \textbf{128\%} (ChartX+A-CoT+SC), \textbf{85\%} (MatPlotBench+A-CoT+SC), \textbf{70\%} (Plot2Code+A-CoT+SC) more tokens on average than \textsc{ChartAF}, they fail to be competitive. At the time of experimentation, an evaluation for a single query costed \$0.1 and \$0.12 on average using \texttt{GPT 4o} and \texttt{Claude 3.5 Sonnet}, respectively.

\subsection{\textsc{ChartUIE-8K} Experiments}\label{sec:uie_exp} 
\paragraph{\textbf{Experiment Set-up.}} We provide evidence than our novel evaluation set is of high-quality, and potentially more realistic than past benchmarks. We employ a double-blind human study of 130 participants (89 of whom pass the rigorous sanity check). Each participant is given a randomized chart image they are imagining, and asked to prompt an LLM their initial instructions. As the factors that comprise query realism are qualitative in nature, we analyze three dimensions that can be quantitatively captured: (i) the word count distribution, (ii) \% of respondants preferring \textsc{ChartUIE-8K}’s interaction, and (iii) \% of respondants who think \textsc{ChartUIE-8K}’s user emulation is realistic. Lastly, the participant is asked whether the extra QA cycle is desirable and realistic when interacting with an LLM. Details of the human study is provided in App. \ref{app:uie_human}.

\begin{figure}
    \centering
\includegraphics[width=1\linewidth]{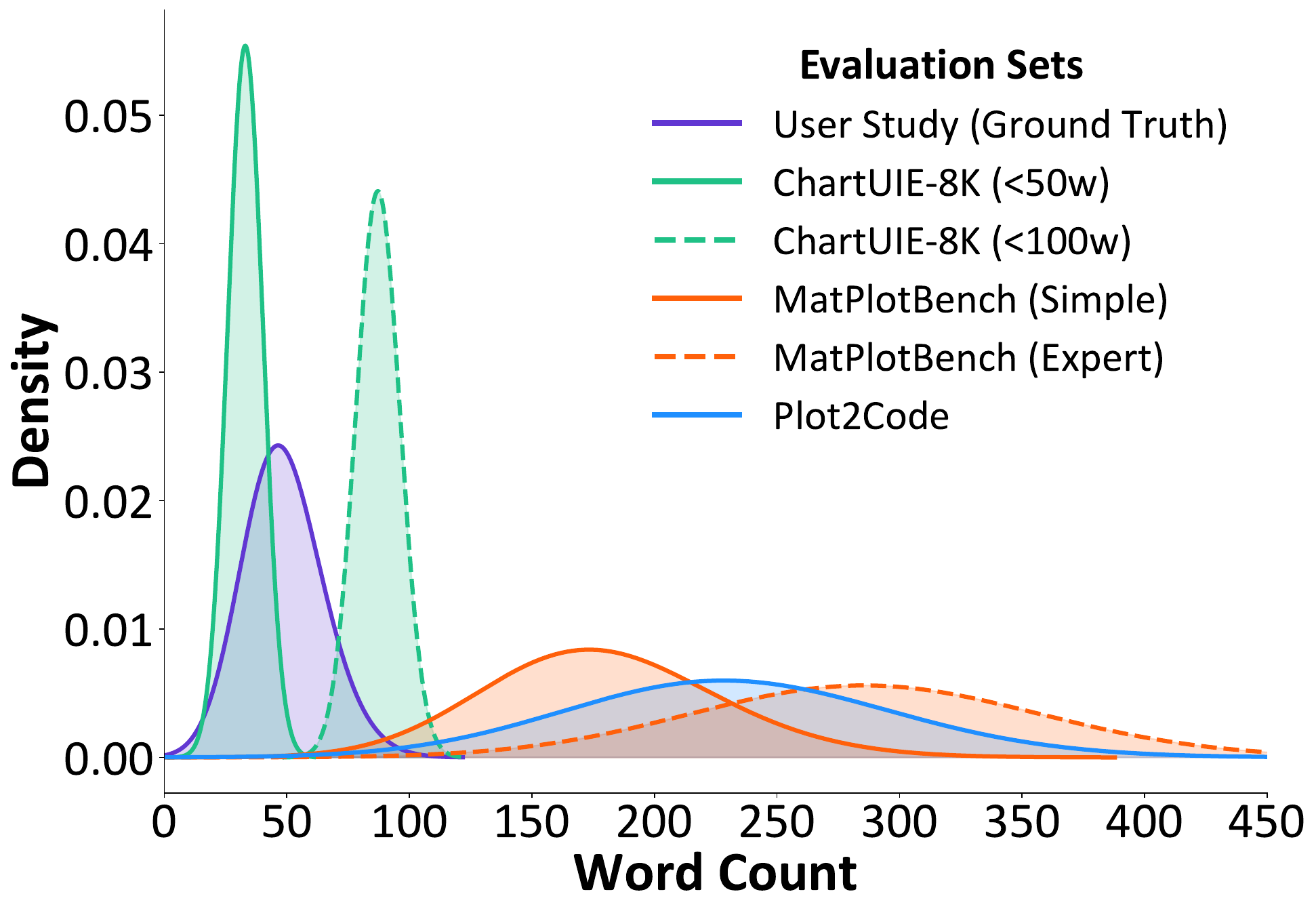}
    \centering
    \includegraphics[width=1\linewidth]{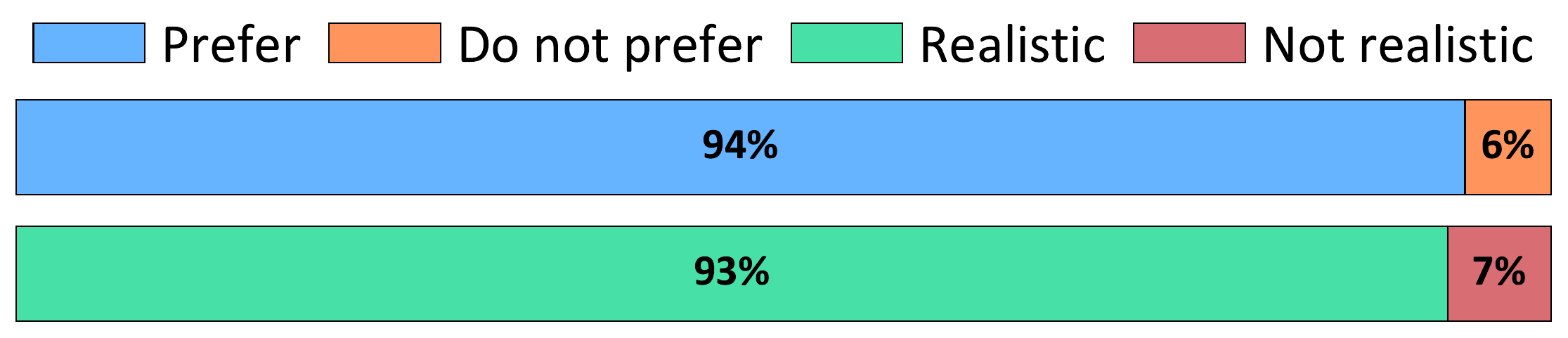}
\caption{Results of the \textsc{ChartUIE-8K} empirical study. \textbf{Top:} Ground truth $\mathcal{Q}^*$ ({\color{user}\textbf{purple}}) vs. $\mathcal{Q}$ distributions of initial instruction word count. \textbf{Middle:} \% of respondants preferring \textsc{ChartUIE}'s interaction. \textbf{Bottom:} \% of respondants who think \textsc{ChartUIE}'s user emulation is realistic.}
    \label{fig:uiestudy}
\end{figure}

\paragraph{\textbf{Results.}} We visualize the results in Fig. \ref{fig:uiestudy}. In terms of word count, \textsc{ChartUIE-8K} ({\color{uie}\textbf{green}}) most closely matches the ground truth distribution of the study of LLM users ({\color{user}\textbf{purple}}). Please see some qualitative examples of the instructions the users gave in App. \ref{sec:user_inst}. Notably, 94\% of respondents prefer the extra QA cycle, and 93\% believe \textsc{ChartUIE}'s user emulation is realistic. We also qualitatively observe, as presented in App. \ref{app:query}, existing evaluation sets' queries are too technical to be realistic. 
\section{Discussion and Impact of \textbf{\textsc{C$^2$}}}\label{sec:dis}
\subsection{\textsc{ChartAF} Enables Scalable Feedback}\label{sec:enable}

\paragraph{Scalability.} As described in Sec. \ref{sec:intro}, the lack of training data, $\langle \text{instruction}, \text{data}, \text{code}\rangle$, can be mitigated by an effective reference-free feedback provider that only requires $\langle \text{instruction}, \text{data}\rangle$. However, as indicated by the predominance of {\color{darkred}\textbf{red}} in Tab. \ref{table:exp1}, existing feedback providers perform poorly under this reference-free regime. Furthermore, increasing token usage via current methods fails to resolve this issue. \textsc{ChartAF} addresses this scalability barrier, as demonstrated by the experimental results in in-context tuning and TTS. 

\paragraph{In-context Tuning with \textsc{ChartAF}.} \textsc{ChartAF} performs best among existing methods for nearly all $f$ and $f_{AF}$ (Tab. \ref{table:exp1}). This is particularly notable, as in-context tuning is often challenging for smaller models. Even when the feedback is helpful, the intrinsic limitations of $f$—such as small model size or insufficient tuning to follow instructions—may result in $f$ inaccurately reflecting the feedback. Despite these challenges, the granular feedback of \textsc{ChartAF} effectively improves smaller models (\texttt{Llama 3.1 70b}, \texttt{Gemma 2 27b}). In fact, \texttt{Llama 3.1 70b} and \texttt{Gemma 2 27b} on average experience greatest performance improvements after in-context tuning with \textsc{ChartAF}, likely due to their weaker base performance that allows for larger post-tuning gains.

\paragraph{Towards Complex Task Verification.} Additionally, \textsc{ChartAF}'s reference-free TTS performance shows significant strength. TTS has been a paradigm-shifting development as it introduces a novel neural scaling axis. While previous scaling laws focused on data and training compute, recent works show that similar scaling laws apply to the inference compute axis. However, current TTS is limited to tasks with reliable and cheap verifiers \citep{wang2024litesearch,wang2024q}, emphasizing the salience of fast, reliable, and cost-effective verifiers \citep{brown2024large,snell2024scaling}. \textsc{ChartAF} is the first effective demonstration of a chart generation verifier within the TTS framework.

We encourage future works to build upon our approach to advance TTS for chart generation and other similar tasks. We used the simplest approach for TTS to empirically prove the verifier's effectiveness. More advanced TTS for chart generation is an open problem.

\paragraph{Scaling Frontier Models with \textsc{ChartAF}.} Notably, \textsc{ChartAF} is not a distillation method transferring knowledge from larger$\rightarrow$smaller models. In Tab. \ref{table:exp1}, \textsc{ChartAF} remains effective even for $\langle f, f_{AF} \rangle$ pairs where $f$ and $f_{AF}$ are similarly performing models. Furthermore, TTS is demonstrated using the same backbone LLM. Such self-improving LLMs are indispensable in advancing frontier models \citep{huang-etal-2023-large}.

\paragraph{Scaling without Parameter Updates.} We demonstrate the effectiveness of \textbf{\textsc{C$^2$}} with no parameter updates. This is notable as parameter updating LLMs incur a large memory and training throughput cost. Therefore, \textbf{\textsc{C$^2$}} is orthogonal to \citet{zadeh2024text2chart31}, where they focus on automating the instruction (parameter) tuning process for chart generations.

\subsection{Unlocking Large-scale Data with \textsc{C$^2$}}

\paragraph{Comprehensive Evaluation and Generation.} The strong reference-free performance of \textsc{ChartAF} (Sec. \ref{sec:enable}) enables the curation of a cost-effective and diverse query set, \textsc{ChartUIE-8K}. This approach contrasts with the reference-based nature of existing query sets, which suffer from limited diversity (Tab. \ref{tab:uie}). By pairing \textsc{ChartAF} with \textsc{ChartUIE-8K}, the combined framework of \textbf{\textsc{C$^2$}} supports a \textit{broadly} inclusive evaluation set (Tab. \ref{tab:uie}, Fig. \ref{fig:dist}). This enables researchers to evaluate models on a more comprehensive and diverse set of queries. Moreover, through \textsc{ChartAF}, \textbf{\textsc{C$^2$}} can generate large-scale, high-quality outputs that significantly improve over previous methods.

\paragraph{Realistic Evaluation.} Lastly, it is crucial to create evaluation sets that closely reflect real-world use-cases. As shown by the distributions in Fig. \ref{fig:uiestudy}, existing query sets often diverge significantly from common user queries, reducing their practical utility. We recommend that future work proposing evaluation sets include rigorous studies (e.g. Sec. \ref{sec:uie_exp}) to ensure their assets are pragmatically aligned with real-world use-cases.



\clearpage
\section{Limitations}
We discuss the limitations of this work to ensure full transparency. To our knowledge, we disclose all reasonable information throughout the paper, project site, and github repository.

\paragraph{Code Execution Error.} The feedbacks, $s_\mathbb{N}$, $s_\mathcal{N}$, presented in this paper is conditioned on the fact that the chart image has been successfully generated from the code. Occasionally, the code fails to execute or executes with an error. Under such circumstances, we allow a maximum of 5 re-generations for the initial $f$ inference, and a maximum of 3 re-generations for post-feedback $f$ inference. We document the initial $f$ inference error rate for each $f$ in App. \ref{app:bench}. 

\paragraph{Coverage.} The coverage of \textsc{ChartUIE-8K} and \textsc{ChartAF} is limited to the English language. Additionally, this study is not conducted with smaller models, e.g. 8B parameter size LLMs. We leave investigating expanded coverage to future works.

\section{Human Study Ethical Consideration}

To our knowledge, we follow  best practices in computer science human studies \citep{muller2014survey,43803}. First, to guarantee the privacy of both researchers and surveyors the study is double-blind, and do not collect any unnecessary data. Second, we clearly indicate at the very start that this is an academic survey for an academic paper. We do not upload the raw data on the public domain to avoid any potential unethical usage. Third, surveyors voluntarily conduct the surveys and can choose to leave at any time. Fourth, we ensure that the surveyors are compensated fairly. While we do not include any geographic restrictions, we pay $\$14$ per hour, well above the U.S. federal minimum wage of $\$7.25$ \citep{henderson2024crisis} as of the surveys. We pay surveyors within 48 hours of completing the survey. Finally, our surveys do not contain explicit or triggering content.

\section*{Acknowledgement}
We would like to thank Yun Ho Ro for his assistance with the figures and Jian Kim for her help in preparing the human study survey.
This work was supported in part by the Yonsei University Research Fund of 2024-22-0058, the National Research Foundation of Korea (NRF) Grant funded by the Korea government (MSIT) (RS-2024-00353125), and the Institute of Information and Communications Technology Planning and Evaluation (IITP) Grant, Artificial Intelligence Graduate School Program, Yonsei University (RS-2020-II201361) and KAIST (RS-2019-II190075).

\clearpage
\bibliography{custom}

\newpage
\appendix

\renewcommand{\contentsname}{\Large Contents}
\tableofcontents

\newpage
\onecolumn
\newcounter{prompt}
\newpage \section{\textsc{ChartAF} Details}\label{app:chartaj}

\subsection{Basic Criteria References}\label{app:ref}
See Table \ref{tab:criteria_ref}.

\begin{table*}[h]
    \centering
  \resizebox{\textwidth}{!}{%
    \begin{tabular}{ll}
    \toprule
        \textbf{Criteria Category} & \textbf{References} \\
    \midrule
        Chart Type & \citet{figueiras2013typology,jung2017chartsense,islam2019overview,midway2020principles}\\
        Visual Embellishment & \citet{bateman2010useful,borgo2012empirical,andry2021interpreting}\\
    Text&\citet{chi2015morphable,stokes2022striking,stokes2023role}\\
        Color &\citet{healey1996choosing,lee2012perceptually,tennekes2014tree,rhyne2017applying}\\
        Annotation & \citet{lee2015more,ren2017chartaccent, lavrivc2017collaborative,chen2018supporting,kim2021towards,zhao2021chartstory}\\
         Aesthetics&\citet{skog2003between,cawthon2007effect,kim2013topics,harrison2015infographic}\\
         Visual Clutter&\citet{rosenholtz2005feature,ellis2007taxonomy,9385921}\\
    \bottomrule
    \end{tabular}}
    \caption{References that form the theoretical background of the Basic Criteria}
    \label{tab:criteria_ref}
\end{table*}

\subsection{\textsc{ChartAF} Pseudocode}
We present the pseudocode of \textsc{ChartAF} in Alg. \ref{alg:chartaj}, along with the prompts used within it. These prompts can be accessed by clicking on the highlighted phrases in the pseudocode. 


We first introduce the notations. Let $inst$ represent the initial instruction, $qst$ the follow-up questions, and $ans$ the answers to $qst$. Let $d$ represent the dataset, and $d_{attr}$ its attributes. $f_{AF}$ denotes the backbone LLM for feedback generation. Let $mode$ be one of two values: "Scalar" or "Granular." Finally, let $code_{gen}$ be the code that generates a chart. This code is produced by the chart-generating LLM, using the prompt \hyperref[prompt:ca]{$Generate$}. The arguments in the prompt—\texttt{data\_path}, \texttt{data}, \texttt{file\_index}, \texttt{initial\_instruction}, \texttt{questions}, and \texttt{answers}—should be set to the data path for the dataset, $d$, the index for the resulting image, $inst$, $qst$, and $ans$, respectively. Denote $img$ as the chart generated by executing $code_{gen}$. The procedure returns either a single scalar value, $s_\mathbb{N}$, or fine-grained natural language feedback, $s_\mathcal{N}$, depending on the value of $mode$. 

\begin{algorithm}[H]
\caption{\label{alg:chartaj}ChartAF} 
	\begin{algorithmic}[1]
\Procedure{ChartAF}{$inst, qst, ans, d_a, d, $ $f_{AF}, mode, code_{gen}, img$}
\State $p_{tpa} \leftarrow $ \hyperref[prompt:aj-tpa]{$TPA_{AF}$}
\State $p_{tpa}$.format(
\State \ \ \ \ \texttt{initial\_instruction}:=$inst$,
\State \ \ \ \ \texttt{tasks}:=\hyperref[tasks]{$task$},
\State \ \ \ \ \texttt{data}:=$d$
\State \ \ )
\State $\langle tsk, prps, aud\rangle \leftarrow f_{AF}(p_{tpa})$
\State $p_{crt} \leftarrow $ \hyperref[prompt:criteria]{$Criteria$}
\State $p_{crt}$.format(
\State \ \ \ \ \texttt{initial\_instruction}:=$inst$,
\State \ \ \ \ \texttt{questions}:=$qst$,
\State \ \ \ \ \texttt{answers}:=$ans$,
\State \ \ \ \ \texttt{tasks}:=$tsk$,
\State \ \ \ \ \texttt{purpose}:=$prps$,
\State \ \ \ \ \texttt{audience}:=$aud$
\State \ \ )
\State $crt \leftarrow f_{AF}(p_{crt})$
\State $p_{qst} \leftarrow $ \hyperref[prompt:question]{$Criteria_Q$}
\algstore{chartaj}
    \end{algorithmic}
\end{algorithm}

\begin{algorithm}[H]
\ContinuedFloat
\caption{\label{alg:chartaj_cont}ChartAF (Continued)} 
    \begin{algorithmic}[1]
\algrestore{chartaj}
\State $p_{qst}$.format(
\State \ \ \ \ \texttt{task}:=$tsk$,
\State \ \ \ \ \texttt{purpose}:=$prps$,
\State \ \ \ \ \texttt{audience}:=$aud$,
\State \ \ \ \ \texttt{criteria}:=$crt$
\State \ \ )
\State $crt_{qst} \leftarrow f_{AF}(p_{qst})$
\State $p_{eval} \leftarrow $ \hyperref[prompt:eval]{$Evaluate$}
\State $p_{eval}$.format(
\State \ \ \ \ \texttt{evaluation\_questions}:=$crt_{qst}$,
\State \ \ \ \ \texttt{initial\_instruction}:=$inst$
\State \ \ )
\State $s \leftarrow f_{AF}(img, p_{eval})$
\State $s_{\mathbb{N}} \leftarrow $ Ratio of "yes" responses in $s$
\If{$mode$ is "Scalar"}
    \State \textbf{return} $s_{\mathbb{N}}$
\EndIf
\State $s_{\mathcal{N}} \leftarrow $ Feedback in $s$
\State $\langle rtn, dsc, edt, add\rangle \leftarrow $ Classification of \\ \ \ \ \ \ \ \ \ \ \ \ \ \ \ \ \ feedback according to the \texttt{tag} in $s_{\mathcal{N}}$
\State $p_{cf} \leftarrow $ \hyperref[prompt:cf]{$CF$}
\State $p_{cf}$.format(
\State \ \ \ \ \texttt{initial\_instruction}:=$inst$,
\State \ \ \ \ \texttt{code}:=$code_{gen}$,
\State \ \ \ \ \texttt{attributes}:=$d_{attr}$,
\State \ \ \ \ \texttt{retain}:=$rtn$,
\State \ \ \ \ \texttt{discard}:=$dsc$,
\State \ \ \ \ \texttt{edit}:=$edt$,
\State \ \ \ \ \texttt{add}:=$add$
\State \ \ )
\State $s_{\mathcal{N}} \leftarrow f_{AF}(p_{cf})$
\State \textbf{return} $s_\mathcal{N}$
\EndProcedure
	\end{algorithmic}
\end{algorithm}

\refstepcounter{prompt}
\begin{tcolorbox}[enhanced jigsaw,breakable,pad at break*=1mm,colback=orange!5!white,colframe=orange!50!black, colbacktitle=orange!75!black,title=Generating a Chart ($Generate$),label=prompt:ca]
You are an expert data visualizer.

The following instruction asks you to generate code for data visualization of the underlying data file that we have attached.
I will give you the data, but you can ignore some parts from the data if it is not necessary and unrelated to the instruction.
Assume that the data file that has been attached in the path “\{data\_path\}” in the generated code.
The file format of the data is **f”.json or .csv”.**
Your code should include loading the data file, and check and verify the data type and representation of the data to avoid errors while executing.
\\

<start of data format>\\
\{\texttt{data}\}\\
<end of data format>
\\

Your code should also automatically download the final visualization in a lower level directory (contained within the current directory) named “plots\_d2c”.
You MUST name your final generated visualization as ”\{\texttt{file\_index}\}.png”.
You can freely choose package(s) that work best to make the visualization.
\\

Here is the instruction set:
\\

<start of initial instruction>\\
\{\texttt{initial\_instruction}\}\\
<end of initial instruction>
\\

<start of further instruction>\\
Questions:\\
\{\texttt{questions}\}\\
Answers:\\
\{\texttt{answers}\}\\
<end of further instruction>
\\

Ensure you use this code format in order to avoid errors, and only give the executable Python Code.
\\







\end{tcolorbox}

\refstepcounter{prompt}
\begin{tcolorbox}[enhanced jigsaw,breakable,pad at break*=1mm,colback=orange!5!white,colframe=orange!50!black, colbacktitle=orange!75!black,title=\textsc{ChartAF} Task\, Purpose\, and Audience Inference ($TPA_{AF}$),label=prompt:aj-tpa]
You are a data visualization expert.
Given the data and user request, your task is to analyze the user request to (1) select the most suitable task that the user is expecting from the list of various tasks in data visualization, (2) specifically figure out the purpose of the user's request in data visualization, and (3) prospective audience of the data visualization.
\\

<start of user request> \\
\{\texttt{initial\_instruction}\} \\
<end of user request>
\\

<start of data format>\\
\{\texttt{data}\}\\
<end of data format>
\\

<start of various task types>\\
\{\texttt{tasks}\}\\
<end of various task types>\\


\end{tcolorbox}

We use data visualization tasks presented in \citet{10.1145/3154862.3154881}.

\refstepcounter{prompt}
\begin{tcolorbox}[enhanced jigsaw,breakable,pad at break*=1mm,colback=orange!5!white,colframe=orange!50!black, colbacktitle=orange!75!black,title=List of Tasks ($task$),label=tasks]
\begin{itemize}
    \item Show External Context \\ Uncaptured data provided by the self-tracker to understand and explain a phenomenon shown in the data.
    \item Show confirmation \\ Collected data confirms existing knowledge.
    \item Show Contradiction \\ Collected data contradicts existing knowledge.
    \item Focus on Identifying value \\ Explicitly specify the measured value, its range for one or more clearly identified data points, or the difference between two measured values.
    \item Focus on Identifying extreme \\ Explicitly state the identities of the data points possessing extreme values of the measure variable.
    \item Focus on Identifying reference \\ Explicitly state the values of categorical variables, labels from the axes, or legends.
    \item Comparison by Time Segmentation \\ Compare measured values segmented by time.
    \item Comparison by Multiple services \\ Compare the same data type from two or more services.
    \item Comparison against external data \\ Bringing in external data for comparison.
    \item Comparison by Factor \\ Compare measured values by a factor (other than time).
    \item Comparison by Instances \\ Compare two specific instances.
    \item Show Trend \\ Describe changes over time.
    \item Value judgement \\ Convey positive or negative connotations about the data.
    \item Distribution with variability \\ Explicitly state the variability of measured values.
    \item Distribution By Category \\
    Explicitly describe the variation of measured values across all or most of the values of a categorical variable.
    \item Correlation \\ Specify the direct relationship between two variables (but not as comparison).
    \item Outlier \\ Explicitly point out outliers or state the effect of outliers.
    \item Sumarization of data \\ Summary of collected data (such as number of data points, duration of tracking, and averages).
    \item Prediction/Forecasting \\ Predict the future based on the collected data.
\end{itemize}
\end{tcolorbox}

\refstepcounter{prompt}
\begin{tcolorbox}[enhanced jigsaw,breakable,pad at break*=1mm,colback=orange!5!white,colframe=orange!50!black, colbacktitle=orange!75!black,title=\textsc{ChartAF} Criteria Establishment ($Criteria$),label=prompt:criteria]
You are a data visualization expert.
You are given basic essential requirements of chart, user instruction, user request with QA, tasks that must be covered by the chart, purpose of the chart, and prospective audience of the chart.

Your task is to develop a personalized, detailed, and objective list of criteria, building on the basic criteria, to evaluate a data visualization (chart).
These criteria should be based on the user instruction, the user request through questions and answers, tasks at hand, intended purpose, and the prospective audience.
\\

<start of basic criteria>
\begin{itemize}
    \item Chart Type \\ Choose a chart type that aligns with the given purpose, task, and audience. The chart type should effectively convey the intended message; for example, bar charts are ideal for comparing quantities for limited number of categorical data, while line charts show trends over time. The choice must consider the inherent spacing requirements and the context in which the chart will be used, ensuring clarity and comprehension.
    \item Visual Embellishment \\ Use embellishments to enhance understanding without overwhelming the data. Visual embellishments, like icons, patterns, or textures, should be used sparingly and purposefully to make the chart memorable and engaging while maintaining a balance that does not distract from the core data.
    \item Text \\ Prioritize legibility and adhere to consistent textual criteria. Text elements, such as legends, titles, and labels, should be legible and easy to read, with sufficient contrast against the background. Consistent font size, style, and placement should be maintained to create a cohesive visual narrative that guides the audience's understanding.
    \item Color \\ Use color purposefully and sparingly to convey meaning. Choose a limited palette that enhances readability and highlights key data points, considering color statistics and opponent processing principles (contrasting colors for clarity). This helps ensure accessibility for viewers with color vision deficiencies.
    \item Annotation \\ Emphasize critical data while minimizing irrelevant details. Use annotations strategically to draw attention to important insights, trends, or outliers, and smooth over or de-emphasize less significant data points, ensuring the chart communicates its key message effectively.
    \item Aesthetics \\ Tailor aesthetics to the chart's purpose, audience, and context. Consider the chart’s purpose, the target audience, and the presentation environment when designing aesthetics, including compact spacing and visual hierarchy. This ensures the chart is both functional and appealing, maximizing its impact and effectiveness
    \item Visual Clutter \\ Optimize the chart size to fit its content and context, balancing data and available space to prevent clutter or excessive white space while maintaining readability. Manage visual elements by minimizing overcrowding and overlapping, adequately spacing text, data points, and annotations, removing unnecessary details, and maintaining a clean layout to enhance clarity. Segmentation of complex charts or data visualizations can also be employed if the visual complexity is high, breaking down the data into smaller, more manageable parts for easier interpretation. It is important to emphasize key data by using size, color, and opacity to highlight critical insights while downplaying less relevant information for a focused presentation.
\end{itemize}
<end of basic criteria>
\\

<start of user instruction>\\
\{\texttt{initial\_instruction}\}\\
<end of user instruction>
\\

<start of user request through QA>\\
\{\texttt{questions}\}\\
\{\texttt{answers}\}\\
<end of user request through QA>
\\

<start of tasks>\\
\{\texttt{tasks}\}\\
<end of tasks>
\\

<start of purpose>\\
\{\texttt{purpose}\}\\
<end of purpose>
\\

<start of prospective audience>\\
\{\texttt{audience}\}\\
<end of prospective audience>
\\

Note that the interactivity of the chart, file format, credibility and integrity of data source, summary statistics do not need to be considered.
The quality of the data visualization to general audience is the only subject to be considered.

Think about the essential chart component requirements that align with the task, purpose, and user request.
\end{tcolorbox}

\refstepcounter{prompt}
\begin{tcolorbox}[enhanced jigsaw,breakable,pad at break*=1mm,colback=orange!5!white,colframe=orange!50!black, colbacktitle=orange!75!black,title=\textsc{ChartAF} Generate Criteria Questions ($Criteria_{Q}$),label=prompt:question]
You are an expert critic.
You will be given wanted tasks, intended purpose, prospective audience, and established criteria for the chart that you gave in the previous prompt.
Your task is to create a list of Yes/No questions that checks if the generated chart satisfies the established criteria.
Use the established criteria as a reference, but avoid applying them directly when crafting questions to evaluate the chart.
\\

<start of task>\\
\{\texttt{task}\}\\
<end of task>
\\

<start of purpose>\\
\{\texttt{purpose}\}\\
<end of purpose>
\\

<start of prospective audience>\\
\{\texttt{audience}\}\\
<end of prospective audience>
\\

<start of established criteria>\\
\{\texttt{criteria}\}\\
<end of established criteria>
\\

"Yes" should be treated as satisfaction, while "No" should be a dissatisfaction.

Here is a detailed protocol for making questions:

First, create questions according to the criteria, tasks, purpose, and audience. Extra questions that the criteria do not cover can be generated, yet it must help judge evaluating the chart.
Lastly, summarize similar questions and rank the questions so that the first question is the most important and the last question is the least important.

Your output should follow the format below:
\\

'''\\
Question 1 : [Question]\\
Question 2 : [Question]\\
...\\
'''
\end{tcolorbox}

\refstepcounter{prompt}
\begin{tcolorbox}[enhanced jigsaw,breakable,pad at break*=1mm,colback=orange!5!white,colframe=orange!50!black, colbacktitle=orange!75!black,title=\textsc{ChartAF} Evaluation ($Evaluate$),label=prompt:eval]
You are an expert evaluator (judge, critic) of the attached data visualization image.
\\

<start of evaluation questions>\\
\{\texttt{evaluation\_questions}\}\\
<end of evaluation questions>
\\

The evaluation questions consist of YES/NO questions; the answer for each question MUST be either YES or NO. Don't give anything else like N/A.
With the answers, you need to give feedback.

When answering the questions, follow the step-by-step protocol below:

\begin{enumerate}
    \item Determine and tag whether the question is subjective or fact-checking
    \begin{itemize}
        \item Fact-checking \\ Verify if the chart image meets the criteria directly based on the visual content. If the image shows any deviation from the criteria, answer NO. If the image meets the criteria, answer YES.
        \item Subjective \\ Consider whether the image meets the criteria based on visual appeal, clarity, and other subjective measures. Provide reasons for both YES and NO answers. If there is clear evidence to support a YES and no substantial reasons to support a NO, answer YES. Answer NO otherwise.
    \end{itemize}
    \item Answer the questions and provide feedback \\
    After answering each question, provide feedback explaining your evaluation.
    List potential improvements categorized as RETAIN, DISCARD, EDIT, or ADD if necessary.

    Feedback Classification:
    \begin{itemize}
        \item RETAIN \\ Identify and specify any elements that should be retained even after the improvement.
        \item DISCARD \\ Identify and specify any elements that should be discarded for better visualization.
        \item EDIT \\ Specify edits needed in the image to satisfy the user's request. Provide examples if applicable.
        \item ADD \\ Identify and specify elements that should be added for better visualization of the user's initial prompt.
    \end{itemize}
\end{enumerate}

To help your task, here is the user's initial prompt.
\\

<start of initial prompt>\\
\{\texttt{initial\_instruction}\}\\
<end of initial prompt>
\\

\end{tcolorbox}

\refstepcounter{prompt}
\begin{tcolorbox}[enhanced jigsaw,breakable,pad at break*=1mm,colback=orange!5!white,colframe=orange!50!black, colbacktitle=orange!75!black,title=\textsc{ChartAF} Generate Code Feedback ($CF$),label=prompt:cf]
You are an expert software engineer on the Quality Assurance team
Your task is to provide feedback on the code based on the critic's feedback on the result of the code.
The code's goal is to successfully draw a chart, fulfilling the user's needs.
You will be given the user's needs, the original code, the critic's feedback, the data attributes, and the resulting image of the code.
\\

Here is the user's needs.\\
<start of needs>\\
\{\texttt{initial\_instruction}\}\\
<end of needs>
\\

Here is the original code.\\
<start of the code>\\
\{\texttt{code}\}\\
<end of the code>
\\

Here are the data attributes.\\
<start of the attributes>\\
\{\texttt{attributes}\}\\
<end of the attributes>
\\

Here is the critic's feedback.\\
<start of feedback>\\
Elements to RETAIN\\
---\\
\{\texttt{retain}\}\\
\\

Elements to DISCARD\\
---\\
\{\texttt{discard}\}\\
\\

Elements to EDIT\\
---\\
\{\texttt{edit}\}\\
\\

Elements to ADD\\
---\\
\{\texttt{add}\}\\
<end of feedback>
\\

Your task is to provide feedback on the code for debugging and offering better data visualization. Specifically, focus on cases where the image does not correctly reflect the intended output, even though the code appears correct. Follow these steps:

\begin{enumerate}
    \item Review the Evaluation Feedback \\ Examine the feedback, especially noting where the image does not align with the expected results despite the code being correct.
    \item Analyze the Feedback \\ Determine what changes are necessary in the code to correct errors and enhance the output based on the feedback.
    If there are potential errors that may occur, feel free to provide feedback on those lines.
    Again, your task is not only to offer better data visualization but also to debug the code.
    \item List your feedback on the code, and make sure such modifications help generate the executable code. \\ Explain the modification, log the lines of code that should be modified, and log lines of new code that can be implemented.
    When logging the code, log the line number as well, where the original code lies, and where the new code should be put.
\end{enumerate}

\end{tcolorbox}

\newpage \section{\textsc{ChartUIE-8K} Details}\label{app:uie}
\subsection{\textsc{ChartUIE-8K} Pseudocode}
We present the pseudocode of \textsc{ChartUIE-8K} in Alg. \ref{alg:uie}, along with the prompts used within it. These prompts can be accessed by clicking on the highlighted phrases in the pseudocode.

We first introduce the notations. Let $d$ represent the underlying dataset. If it exists, let $d_{ttl}$ denote the title of $d$; otherwise, set $d_{ttl}$ to "unknown." Similarly, if the topic of $d$ is provided, let $d_{tpc}$ represent it. Finally, let $f_{uie}$ refer to the backbone LLM for \textsc{ChartUIE-8K}. Here, $f_{uie}:=$\texttt{GPT 4o}.

The procedure returns a list of queries, $Q$, generated by the algorithm. Each tuple in $Q$ includes the initial instruction ($inst$), column labels for visualization (data attributes, $d_{attr}$), a selected task ($tsk$), a visualization purpose ($prps$), a text description of the target audience ($aud$), follow-up questions ($qst$), and answers ($ans$). The model $f_{uie}$ infers $d_{attr}$, $tsk$, $prps$, and $aud$ from $inst$ and follow-up questions to clarify user preferences, of which only $q_{\%}$ are answered.

\label{app:uie_pseudo}
\begin{algorithm}[H]
\caption{\label{alg:uie}\textsc{ChartUIE-8K}} 
	\begin{algorithmic}[1]
\Procedure{UIE-8K}{$d, d_{ttl}, d_{tpc}, f_{uie}$}
\State $p_{ct} \leftarrow $ \hyperref[prompt:ct]{$Select_{ct}$}
\State $p_{ct}$.format(\texttt{data}:$=d$)
\State $C_{type} \leftarrow f_{uie}(p_{ct})$
\State $p_{annot} \leftarrow $ \hyperref[prompt:an]{$Select_{annot}$}
\State $p_{annot}$.format(
\State \ \ \ \ \texttt{data}:=$d$,
\State \ \ \ \ \texttt{data\_title}:=$d_{ttl}$,
\State \ \ \ \ \texttt{topic}:=$d_{tpc}$
\State \ \ )
\State $annot \leftarrow f_{uie}(p_{annot})$
\State $j_{max} \leftarrow \min \{15, \text{the length of } C_{type}\}$
\State $Q \leftarrow []$
\For{$j \in \{0, 1, ..., j_{max} - 1\}$}
    \For{$word \in \{50, 100\}$}
        \If{$j \leq 2$}
            \State $p_{trg} \leftarrow $ \hyperref[prompt:ian]{$Trigger_{annot}$}
            \State $p_{trg}$.format(
            \State \ \ \ \ \texttt{data}:=$d$,
            \State \ \ \ \ \texttt{annotations}:=$annot$,
            \State \ \ \ \ \texttt{data\_title}:=$d_{ttl}$,
            \State \ \ \ \ \texttt{topic}:=$d_{tpc}$,
            \State \ \ \ \ \texttt{chart\_type}:=$C_{type}[j]$
            \State \ \ \ \ \texttt{word\_count}:=$word$
            \State \ \ )
    \algstore{chartuie}
	\end{algorithmic}
\end{algorithm}

\begin{algorithm}[H]
\setlength{\baselineskip}{0.95\baselineskip} 
\ContinuedFloat
\caption{\label{alg:chartuie}\textsc{ChartUIE-8K} (Continued)} 
    \begin{algorithmic}[1]
\algrestore{chartuie}
        \Else
            \State $p_{trg} \leftarrow $ \hyperref[prompt:ian0]{$Trigger_{annot'}$}
            \State $p_{trg}$.format(
            \State \ \ \ \ \texttt{data}:=$d$,
            \State \ \ \ \ \texttt{data\_title}:=$d_{ttl}$,
            \State \ \ \ \ \texttt{topic}:=$d_{tpc}$,
            \State \ \ \ \ \texttt{chart\_type}:=$C_{type}[j]$
            \State \ \ \ \ \texttt{word\_count}:=$word$
            \State \ \ )
        \EndIf
        \State $inst \leftarrow f_{uie}(p_{trg})$
        \State $p_{tpa} \leftarrow $ \hyperref[prompt:ca-tpa]{$TPA_{uie}$}
        \State $p_{tpa}$.format(
        \State \ \ \ \ \texttt{initial\_instruction}:=$inst$,
        \State \ \ \ \ \texttt{data}:=$d$,
        \State \ \ \ \ \texttt{task}:=\hyperref[tasks]{$task$}
        \State \ \ )
        \State $\langle d_a, tsk, prps, aud \rangle \leftarrow f_{uie}(p_{tpa})$
        \State $p_{qst} \leftarrow $ \hyperref[prompt:ca-q]{$Question_{uie}$}
        \State $p_{qst}$.format(
        \State \ \ \ \ \texttt{data}:=$d$,
        \State \ \ \ \ \texttt{initial\_instruction}:=$inst$,
        \State \ \ \ \ \texttt{attributes}:=$d_{attr}$,
        \State \ \ \ \ \texttt{audience}:=$aud$,
        \State \ \ \ \ \texttt{tasks}:=$tsk$,
        \State \ \ \ \ \texttt{purpose}:=$prps$
        \State \ \ )
        \State $qst \leftarrow f_{uie}(p_{qst})$
        \State $p_{ans} \leftarrow $ \hyperref[prompt:uie-a]{$Answer_{uie}$}
        \If{$word == 50$}
            \State $q_{\%} \leftarrow 25$
        \Else
            \State $q_{\%} \leftarrow 50$
        \EndIf
        \State $p_{ans}$.format(
        \State \ \ \ \ \texttt{initial\_instruction}:=$inst$,
        \State \ \ \ \ \texttt{q\_percent}:=$q_{\%}$,
        \State \ \ \ \ \texttt{purpose}:=$prps$,
        \State \ \ \ \ $f$\_\texttt{response}:$=qst$
        \State \ \ )
        \State $ans \leftarrow f_{uie}(p_{ans})$
        \State $Q.append($
        \State \ \ \ \ $\langle$
        \State \ \ \ \ \ \ $inst, d_a, tsk, prps,$
        \State \ \ \ \ \ \ $aud, qst, ans$
        \State \ \ \ \ $\rangle$
        \State \ \ $)$
    \EndFor
\EndFor
\State \textbf{return} $Q$
\EndProcedure
	\end{algorithmic}
\end{algorithm}

\newpage
We use chart types presented in \citet{2022types}.

\refstepcounter{prompt}
\begin{tcolorbox}[enhanced jigsaw,breakable,pad at break*=1mm,colback=darkblue!5!white,colframe=darkblue!50!black, colbacktitle=darkblue!75!black,title=Chart Type Selection ($Select_{ct}$),label=prompt:ct]
<start of data example format>\\
\{\texttt{data}\}\\
<end of data example format>
\\

Look at the following list and select one or more chart types appropriate for the data. Try to choose as many different charts as possible. Consider the purpose and chart type that can visualize the data well. 

But Chart types that are not reasonable for data visualization of the data example format I attached must be excluded. Respond with the chart types that are compatible with the data.
Please include only one chart type that is similar to each other. Comparison, Correlation, Part-to-whole \& hierarchical, Data over time (temporal), Distribution, Geospatial \& other are the purposes, and the chart types below are suitable for the above purpose.

Look closely at the characteristics of the data, and the annotation should be one that produces as little clutter as possible.
\\

<start of the chart type list> 

1. Comparison
\begin{itemize}[itemsep=-1mm]
\item Bar chart
\item Column chart
\item Grouped bar/column chart
\item Lollipop chart
\item Bullet chart
\item Dot plot
\item Dumbbell
\item Pictogram
\item Icon chart
\item Range chart
\item Radial bar chart
\item Parallel coordinates
\item Radar chart
\item Nightingale chart
\item Waterfall chart
\item Matrix chart
\item Small multiples
\item Word cloud
\item Slope chart
\item Table chart
\item Categorical scatter plot
\item Quadrant chart
\end{itemize}

2. Correlation
\begin{itemize}[itemsep=-1mm]
    \item Heatmap
    \item Bubble chart
    \item Scatter plot
    \item Connected scatter plot
    \item Hexagonal binning
    \item Contour plot
\end{itemize}

3. Part-to-whole \& hierarchical
\begin{itemize}[itemsep=-1mm]
    \item Stacked bar/column chart
    \item Diverging bar chart
    \item Population pyramid
    \item Icon array
    \item Waffle chart
    \item Pie chart
    \item Donut chart
    \item Semi-circle donut chart
    \item Marimekko chart
    \item Treemap
    \item Circular treemap
    \item Convex treemap
    \item Dendrogram
    \item Venn diagram
    \item Euler diagram
    \item Circular gauge
    \item Sunburst chart
    \item Funnel \& pyramid chart
\end{itemize}

4. Data over time (temporal)
\begin{itemize}[itemsep=-1mm]
    \item Area chart
    \item Stacked area chart
    \item Stream graph
    \item Bump chart
    \item Bump area chart
    \item Line chart
    \item Spline chart
    \item Step line chart
    \item Candlestick chart
    \item Gantt chart
    \item Barcode chart
    \item OHLC chart
\end{itemize}

5. Distribution
\begin{itemize}[itemsep=-1mm]
    \item Density plot
    \item Ridgeline plot
    \item Horizon chart
    \item Histogram
    \item Radial histogram
    \item Strip plot
    \item Jitter plot
    \item One-dimensional heatmap
    \item Beeswarm chart
    \item Box chart
    \item Violin plot
\end{itemize}

6. Geospatial \& other
\begin{itemize}[itemsep=-1mm]
    \item Geographic heatmap
    \item Choropleth map
    \item Tile map
    \item Chord diagram
    \item Arc diagram
    \item Sankey
    \item Network diagram
    \item Flowchart
\end{itemize}

<end of the chart type list> 
\\

Your response should ONLY contain the chart types.
Do not include anything else.
\end{tcolorbox}

\refstepcounter{prompt}
\begin{tcolorbox}[enhanced jigsaw,breakable,pad at break*=1mm,colback=darkblue!5!white,colframe=darkblue!50!black, colbacktitle=darkblue!75!black,title=Annotation Selection ($Select_{annot}$),label=prompt:an]
<start of data example format>\\
\{\texttt{data}\}\\
<end of data example format>
\\

<start of data details format>\\
\{\texttt{data\_title}\}\\
\{\texttt{topic}\}\\
<end of data details format>
\\

Look at the following list and select one or more annotations appropriate for the data. Choose two annotations.

<start of the annotation list>
\begin{enumerate}
    \item Text Annotations: \\ 
    Description: Data-driven text annotations display values linked to chart elements, such as data points in a scatterplot. They draw attention to specific elements by highlighting their values.
    Purpose: When only some elements are annotated, the intent is to focus the viewer’s attention on those before examining others.
    Other Uses: Non-data-driven annotations can provide context, orientation, or editorial comments.
    \item Shapes: \\
    Description: Shape annotations include lines, arrows, rectangles, and other shapes. They can highlight or enclose specific chart elements to emphasize or compare them.
    Data-Driven Use: Some shapes, like trend lines, are calculated from the underlying data.
    \item Highlights: \\
    Description: Highlights modify the appearance of chart elements (e.g., size, color) to emphasize or reduce their importance.
    Purpose: Used to distinguish certain elements from others, making them stand out visually
\end{enumerate}
<end of the annotation list> 
\\

Look closely at the characteristics of the data, and the annotation should be one that produces as little clutter as possible.
Also, refer to the data details to create as practical and realistic instructions as possible.
Your response should ONLY contain the annotations, description, purpose, and uses.
Do not include anything else.
\end{tcolorbox}

\refstepcounter{prompt}
\begin{tcolorbox}[enhanced jigsaw,breakable,pad at break*=1mm,colback=darkblue!5!white,colframe=darkblue!50!black, colbacktitle=darkblue!75!black,title=Emulating with annotations ($Trigger_{annot}$),label=prompt:ian]
You are an expert user emulator.
\\

<start of data example format>\\
\{\texttt{data}\}\\
<end of data example format>
\\

<start of annotations format>\\
\{\texttt{annotations}\}\\
<end of annotations format>
\\

<start of data details format>\\
\{\texttt{data\_title}\}\\
\{\texttt{topic}\}\\
<end of data details format>
\\

Given a data format, imagine a chart that visualizes this data as the final output you want from the service provider.
It MUST be a chart that can be created using only data columns.
Consider what purpose the data has and the practical purpose of visualization and include it in the instructions.
You need to imagine a chart with \{chart\_type\} and given annotations that utilizes the data format.
If there are multiple given data formats, imagine a chart with \{\texttt{chart\_type}\} and given annotations that utilizes all the data formats.
Since you are an amateur user, your instruction will be partially SUBJECTIVE and NOT DETAILED.
Also, refer to the data details to create as practical and realistic instructions as possible.
Instructions must reflect the context of the data. 
To emulate a real-world user your instruction should be \{word\_count\} in size (word count). Do not include data path in the instruction.
Your response should ONLY contain the user emulated instruction.
Do not include anything else.
\end{tcolorbox}

\refstepcounter{prompt}
\begin{tcolorbox}[enhanced jigsaw,breakable,pad at break*=1mm,colback=darkblue!5!white,colframe=darkblue!50!black, colbacktitle=darkblue!75!black,title=Emulating without annotations ($Trigger_{annot'}$),label=prompt:ian0]
You are an expert user emulator.
\\

<start of data example format>\\
\{\texttt{data}\}\\
<end of data example format>
\\

<start of data details format>\\
\{\texttt{data\_title}\}\\
\{\texttt{topic}\}\\
<end of data details format>
\\

Given a data format, imagine a chart that visualizes this data as the final output you want from the service provider.
It MUST be a chart that can be created using only data columns.
You need to imagine a chart with \{\texttt{chart\_type}\} that utilizes the data format.
If there are multiple given data formats, imagine a chart with \{chart\_type\} that utilizes all the data formats.
Since you are an amateur user, your instruction will be partially SUBJECTIVE and NOT DETAILED.
Also, refer to the data details to create as practical and realistic instructions as possible.
Instructions must reflect the context of the data.
To emulate a real-world user your instruction should be \{word\_count\} in size (word count).
Your response should ONLY contain the user emulated instruction.
Do not include anything else.
\end{tcolorbox}

\refstepcounter{prompt}
\begin{tcolorbox}[enhanced jigsaw,breakable,pad at break*=1mm,colback=darkblue!5!white,colframe=darkblue!50!black, colbacktitle=darkblue!75!black,title=Task Purpose and Audience Inference for UIE ($TPA_{uie}$),label=prompt:ca-tpa]
You are a data visualization expert.
Given the data and user request, your task is to analyze the user request to figure out the (1) data attributes needed for data visualization, (2) select the most suitable task that the user is expecting from the list of various tasks in data visualization, (3) specifically figure out the purpose of the user's request in data visualization, and (4) prospective audience of the data visualization.\\

<start of user request>\\
\{\texttt{initial\_instruction}\}\\
<end of user request>
\\

<start of data format>\\
\{\texttt{data}\}\\
<end of data format>
\\

<start of various task types>\\
\{\texttt{task}\}\\
<end of various task types>
\\

For data attributes needed in data visualization, store them in query['Data Attribute']. Data attributes MUST match exactly the column names of the data.
Store the selected task from the task types in query['Task'].
Store the purpose of visualization in query['Purpose'].
Store the prospective audience in query['Audience'].

Please reply in the same format without altering the key value.\\
\{"Data Attribute": None, "Task": None, "Purpose": None, "Audience": None\}\\
But, please make sure there is no ' in each keys and values. Use only " for the response. But when you write a value sentence or each data attribute's title, you only can use '. Unless you are writing a sentence or each data attribute's title, you should never include ' in response. If there are multiple pieces of data, there is no need to reveal which file each is. Please consider that JSON conversion must be done properly.
\end{tcolorbox}

\refstepcounter{prompt}
\begin{tcolorbox}[enhanced jigsaw,breakable,pad at break*=1mm,colback=darkblue!5!white,colframe=darkblue!50!black, colbacktitle=darkblue!75!black,title=List Preference Questions ($Question_{uie}$),label=prompt:ca-q]
You are an expert data visualization analyst.
Given data, data attribute, tasks, prospective audience, purposes of the chart(data visualization) from the initial instruction (user request) of the user, you have a 2-step task to do.

(1) First, your task is to figure out the essential chart attribute requirements that the chart must have in order to satisfy such tasks and purposes
(2) Then, create a list of questions to the user if the user have specific chart attribute preference for effective data visualization.

Do NOT include anything else in your response other than the list of questions. Your questions should be primarily focused on retrieving the user's preferences.
Do NOT include any questions related to (1) interactivity of the chart, and (2) the file format of the chart.
\\

<start of data format>\\
\{\texttt{data}\}\\
<end of data format>
\\

<start of data visualization instruction>\\
\{\texttt{initial\_instruction}\}\\
<end of data visualization instruction>
\\

<start of data attributes>\\
\{\texttt{attributes}\}\\
<end of data attributes>
\\

<start of prospective audience>\\
\{\texttt{audience}\}\\
<end of prospective audience>
\\

<start of type of tasks>\\
\{\texttt{tasks}\}\\
<end of type of tasks>
\\

<start of purpose>\\
\{\texttt{purpose}\}\\
<end of purpose>
\end{tcolorbox}

\refstepcounter{prompt}
\begin{tcolorbox}[enhanced jigsaw,breakable,pad at break*=1mm,colback=darkblue!5!white,colframe=darkblue!50!black, colbacktitle=darkblue!75!black,title=Answer Preference Questions ($Answer_{uie}$),label=prompt:uie-a]
\refstepcounter{prompt} 
You are an expert user emulator.
\\

<start of user request>\\
\{\texttt{initial\_instruction}\}\\
<end of user request>
\\

Imagine the best result that can be achieved based on the given instructions and is the final output you want from the service provider with your purpose given below.
Based on your imagination you are to respond to the service provider.
Since you are an amateur user you should concretely answer \{\texttt{q\_percent}\}\% of the questions you think are most important, and the remaining questions should be answer with uncertainty, e.g. “I am not sure”, “I do not know”, "I have no specific preference".
Finally, remember that you are looking for an image, not an interactive data visualization.
\\

<start of purpose>\\
\{\texttt{purpose}\}\\
<end of purpose>
\\

<start service provider message>\\
\{\textit{f}\_\texttt{response}\}\\
<end service provider message>
\\

Also, remember that the service provider can not see the image you have access to.
Your response should ONLY contain the user emulated response.
Do not include anything else.
\end{tcolorbox}

\subsection{Example User Study Initial Instructions}\label{sec:user_inst}

We provide user study initial instruction examples curated by the LLM user participants below.

\begin{tcolorbox}[enhanced jigsaw,breakable,pad at break*=1mm,colback=yellow!5!white,colframe=yellow!50!black, colbacktitle=yellow!75!black,title= User Initial Instruction Example 1]
Please make a bar chart of R\&D investment (in billions of USD) according to country. Make each bar green, except for Japan and India, which you should make blue. Title the chart, "R\&D Investment by Country." Include a red line that connects the maximum of each bar. And add a blue dash line where the average is across the whole chart (including the average value in Billions out to two decimal places).
\end{tcolorbox}

\begin{tcolorbox}[enhanced jigsaw,breakable,pad at break*=1mm,colback=yellow!5!white,colframe=yellow!50!black, colbacktitle=yellow!75!black,title= User Initial Instruction Example 2]
Chart grades over 12 months. Have different colors for each student and label by names. Y axis will be percentage and x axis months. Title will be Students Grades Over the Year.

\end{tcolorbox}

\newpage \section{Chart Generation User Query Examples}\label{app:query}

\begin{tcolorbox}[enhanced jigsaw,breakable,pad at break*=1mm,colback=darkgreen!5!white,colframe=darkgreen!50!black, colbacktitle=darkgreen!75!black,title=Plot2Code Example]

The figure generated by the provided Python code consists of four subplots, each displaying a different transformation of the same image. The image is created using a mathematical function that generates a grid of values. 

The first subplot shows the image rotated by 30 degrees. The second subplot shows the image skewed by 30 degrees along the x-axis and 15 degrees along the y-axis. The third subplot shows the image reflected along the x-axis and scaled by 0.5 along the y-axis. The fourth subplot combines all these transformations: rotation, skewing, scaling, and translation. 

Each subplot has a yellow dashed line indicating the intended extent of the image. The x-axis limits are set from -5 to 5, and the y-axis limits are set from -4 to 4 for all subplots. 

To recreate this figure, you would need to generate the same image and apply the same transformations in each subplot. The transformations are applied using an affine transformation matrix, which allows for rotation, skewing, scaling, and translation. The specific parameters for each transformation are mentioned above. 

The figure is displayed using the matplotlib library in Python, which would need to be installed and imported to recreate the figure. The numpy library is also used to generate the image and would need to be installed and imported as well. 

The image is generated using a mathematical function that creates a grid of values between -3.0 and 3.0 with a step of 0.25. This function uses the numpy exp function to calculate the exponential of the negative square of each value in the grid. The result is a 2D array of values, which is used as the image in the subplots.

The transformations applied to the image in each subplot are done using the Affine2D function from the matplotlib.transforms module. This function creates a 2D affine transformation matrix that can be used to apply various transformations to the image. The specific transformations and their parameters are as follows:

\begin{itemize}
    \item Rotation: 30 degrees
    \item Skew: 30 degrees along the x-axis and 15 degrees along the y-axis
    \item Scale: -1 along the x-axis (which reflects the image) and 0.5 along the y-axis
    \item Translation: 0.5 along the x-axis and -1 along the y-axis
\end{itemize}

The extent of the image in each subplot is set to [-2, 4, -3, 2], and the image is clipped to these bounds. The yellow dashed line indicating the intended extent of the image is drawn using the plot function with these bounds.

Finally, the figure is displayed using the show function from the matplotlib.pyplot module.
\end{tcolorbox}

\begin{tcolorbox}[enhanced jigsaw,breakable,pad at break*=1mm,colback=darkgreen!5!white,colframe=darkgreen!50!black, colbacktitle=darkgreen!75!black,title=MatPlotBench's Simple Instruction Example]
Create a Python script using matplotlib to generate a specific plot with the following detailed parameters:

Initialize a figure with a custom size of 7.5 by 7.5 inches.  
Add a single axis to the figure with a custom aspect ratio and specified position.  
Define X as a linear space from 0.5 to 3.5 with 100 elements.  
Calculate Y1 as 3 plus the cosine of X, Y2 as 1 plus the cosine of 1+X/0.75 divided by 2, and Y3 as random values uniformly distributed between Y1 and Y2.

Set major and minor locators for both x and y axes with major intervals of 1 and minor intervals of 4.  
Set minor formatter for the x-axis to display values with two decimal places.  
Limit the x and y axes to a range from 0 to 4.

For major ticks, set the width to 1.0, length to 10, and label size to 14.  
For minor ticks, set the width to 1.0, length to 5, label size to 10, and label color to '0.25'.  
Add a grid with these specific attributes:  
\begin{itemize}
    \item Linestyle set to "--" (dashed).
    \item Linewidth of 0.5.
    \item Color set to '.25' (a shade of gray).
    \item Z-order set to -10.
\end{itemize}

Plot three lines with distinct characteristics:  
\begin{itemize}
    \item The first line (Blue signal) should use color 'C0', linewidth of 2.5, and be placed at z-order 10.
    \item The second line (Orange signal) should use color 'C1' and linewidth of 2.5.
    \item The third line should consist of scatter markers at every third point, with no linewidth, markersize of 9, marker style 's' (square), marker face color 'none', marker edge color 'C4', and marker edge width of 2.5.
\end{itemize}

Set the title "Anatomy of a figure" and axis labels "x Axis label" and "y Axis label" with specific font sizes:  
\begin{itemize}
    \item Title font size should be 20.
    \item Axis label font sizes should be 14.
\end{itemize}

Add a legend with these specifications:  
\begin{itemize}
    \item Positioned at the "upper right".
    \item Font size set to 14.
\end{itemize}

Annotate the figure, including tick labels, axes labels, grid, etc., using circles, text, and code snippets at specified coordinates.  
The circles should have a radius of 0.15, a border color defined by the royal\_blue variable with an alpha of 0.6, and a white, non-filled center.  
Text annotations should be in both royal blue and black colors, with bold and italic styles.
\end{tcolorbox}

\begin{tcolorbox}[enhanced jigsaw,breakable,pad at break*=1mm,colback=darkgreen!5!white,colframe=darkgreen!50!black, colbacktitle=darkgreen!75!black,title=\textsc{ChartUIE-8K} Example]
Create a bar chart showing the GDP compositions of agriculture,
  industry, and services for the top 5 countries by GDP. Highlight the country with
  the highest percentage of each sector. Add annotations to display exact values for
  each bar segment, focusing particularly on the leading country in each sector.
\end{tcolorbox}

\newpage \section{Human Study Details}\label{app:humanstudy}

\subsection{Compensation and Qualification} We use paid crowdsource surveyors to conduct our human study. We use the same platform as \citet{gudibande2024the}. To ensure quality responses, all voluntary participants are paid $\$14$ per hour, well above the U.S. federal minimum wage of $\$7.25$ \citep{henderson2024crisis}. We ensure that surveyors feel safe to express freely by clearly indicating: "feel free to be wholly honest". Moreover, we enforce five requirements to participate in our study: \textbf{(i)} Masters\footnote{Masters are "Workers who have demonstrated excellence across a wide range of tasks are awarded the Masters Qualification. Masters must continue to pass our statistical monitoring to retain the Masters Qualification."}, \textbf{(ii)} $>80\%$ Approval Rate, \textbf{(iii)} $>50$ Approved, \textbf{(iv)} Casual experience using LLMs, \textbf{(v)} Survey with a large screen ($\geq 10$ inches, $\geq 25.4$ cm). For privacy, we do not gather any other unnecessary data.

\subsection{Sanity Check} 
Following \citet{10.1145/3544548.3581111} and \citet{10.1145/3613904.3642103}, we include multiple sanity check questions within the survey for reliability. These sanity checks are solidified \textit{a priori}, and are \textit{never} changed.
\paragraph{\textsc{ChartUIE-8K}.}
In \textsc{ChartUIE-8K}, we manually remove LLM initial query that clearly suggests that the surveyor did not understand the survey instructions or put in virtually zero time. For transparency, we report \textit{all} the sanity check fail cases on our open-source \href{https://github.com/chartsquared/C-2}{github repository}.   

\paragraph{\textsc{ChartAF}.} For the \textsc{ChartAF} human study, two sanity checks filter very poor quality surveyor responses. First, we remove surveys that were completed in 50\% of the lower-bound completion time. We time the duration it takes to complete a survey, and create a lower- and upper-bound to signal to potential surveys how long our study takes. For example, 10 to 16 minutes. In this case, surveys that were completed within 5 minutes would be removed as this would indicate that \textit{very low effort} was put into the survey. Second, a sanity check question was included, presenting an image option intentionally unrelated to the user instructions. Survey responses that prefer (or strongly prefer) this option is removed as it indicates that the surveyor does \textit{not} understand the task or is putting in very little effort.

\subsection{\textsc{ChartAF} Human Study}\label{app:exp}
\begin{figure*} [h]
    \centering
    \includegraphics[height=0.9\textheight]{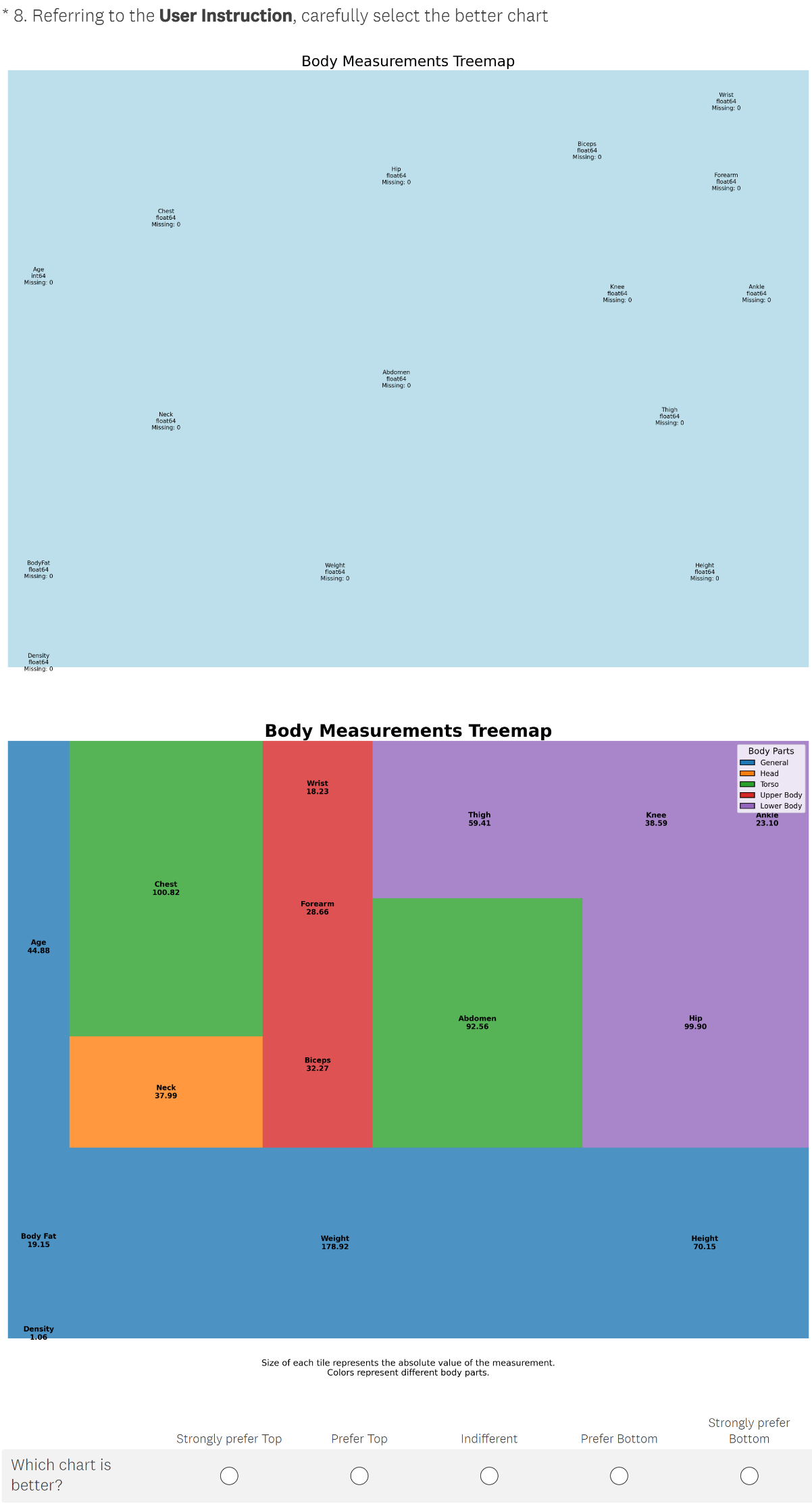}
    \captionsetup{justification=centering}
    \caption{ChartAF Human Study (Part 1) Example}
    \label{fig:pref_1}
\end{figure*}

\begin{figure*} [h]
    \centering
\includegraphics[height=0.9\textheight]{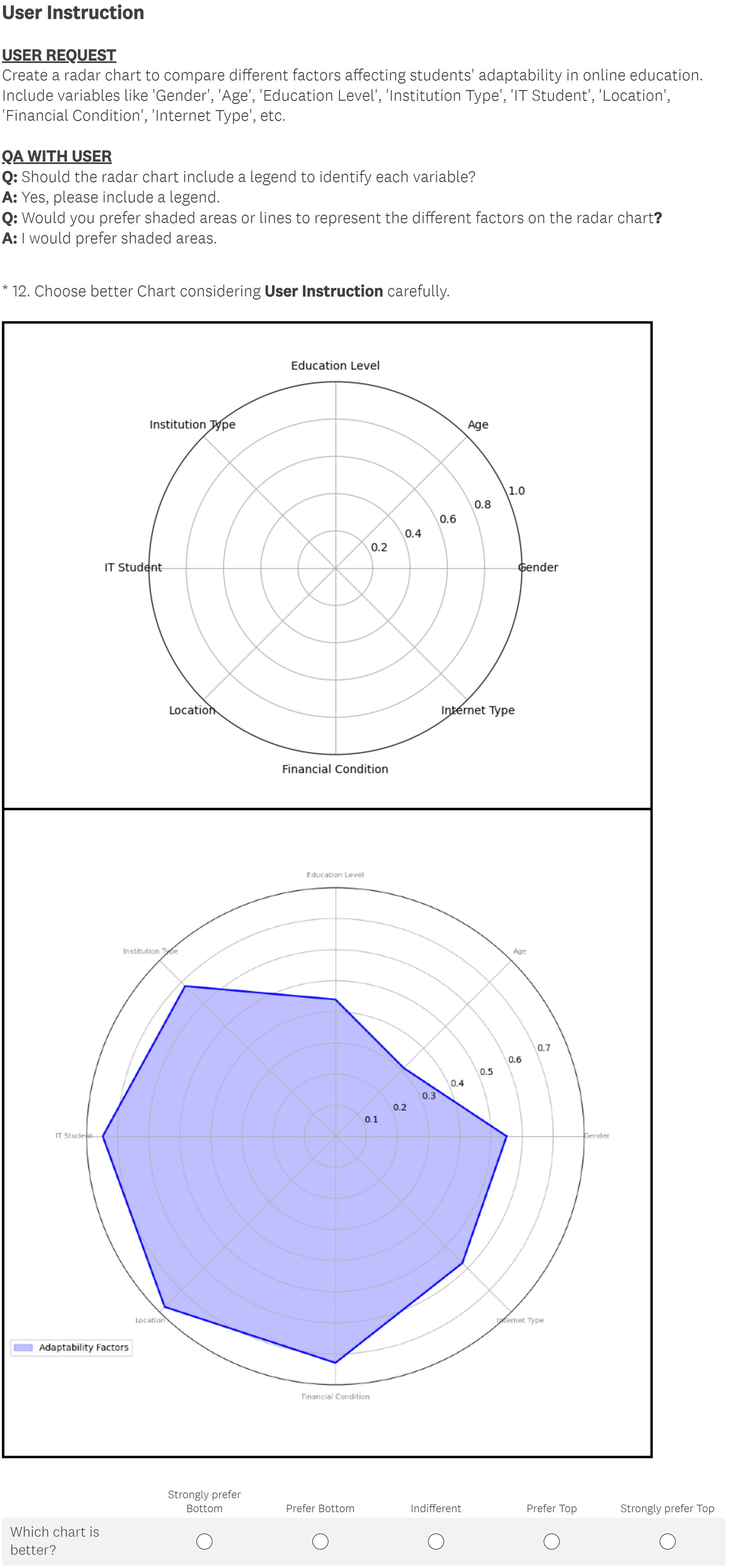}
    \captionsetup{justification=centering}
    \caption{ChartAF Human Study (Part 2) Example}
    \label{fig:pref_2}
\end{figure*}

We provide the precise instructions provided to the surveyors below. We further provide exact screenshots in Fig. \ref{fig:pref_1} and \ref{fig:pref_2}.

\begin{tcolorbox}[enhanced jigsaw,breakable,pad at break*=1mm,colback=black!5!white,colframe=black!50!black, colbacktitle=black!75!black,title=\textsc{ChartAF} Human Study (Part 1)]
\textbf{User Instruction}

\textbf{\underline{USER REQUEST}}

\textit{[insert initial instruction]}

\textbf{\underline{QA WITH USER}}

\textit{[insert QA]}

Referring to the User Instruction, carefully select the \textbf{better} chart.


\textit{[insert generated image by $f$]}

\textit{[insert generated image by \\post-\textsc{ChartAF}-feedback \\or post-ChartX-feedback \\or post-ChartX+A-CoT+SC-feedback \\or Sanity Check \\or post-ChatEval-feedback \\or post-NF-feedback \\or post-MatPlotBench-feedback \\or post-MatPlotBench+A-CoT+SC-feedback \\or post-Plot2Code-feedback \\or post-Plot2Code+A-CoT+SC-feedback \\or post-Skip+A-CoT-feedback]}

\textit{[insert preference score options]}

\end{tcolorbox}

\begin{tcolorbox}[enhanced jigsaw,breakable,pad at break*=1mm,colback=black!5!white,colframe=black!50!black, colbacktitle=black!75!black,title=\textsc{ChartAF} Human Study (Part 2)]
\textbf{User Instruction}

\textbf{\underline{USER REQUEST}}

\textit{[insert initial instruction]}

\textbf{\underline{QA WITH USER}}

\textit{[insert QA]}

Referring to the \textbf{User Instruction}, carefully select the better chart.

\textit{[insert generated image by pre-TTS]}

\textit{[insert generated image by post-TTS]}

\textit{[insert preference score options]}

\end{tcolorbox}

\subsection{\textsc{ChartUIE-8K} Human Study}\label{app:uie_human}
\begin{figure*} [h]
    \centering
\includegraphics[width=\textwidth]{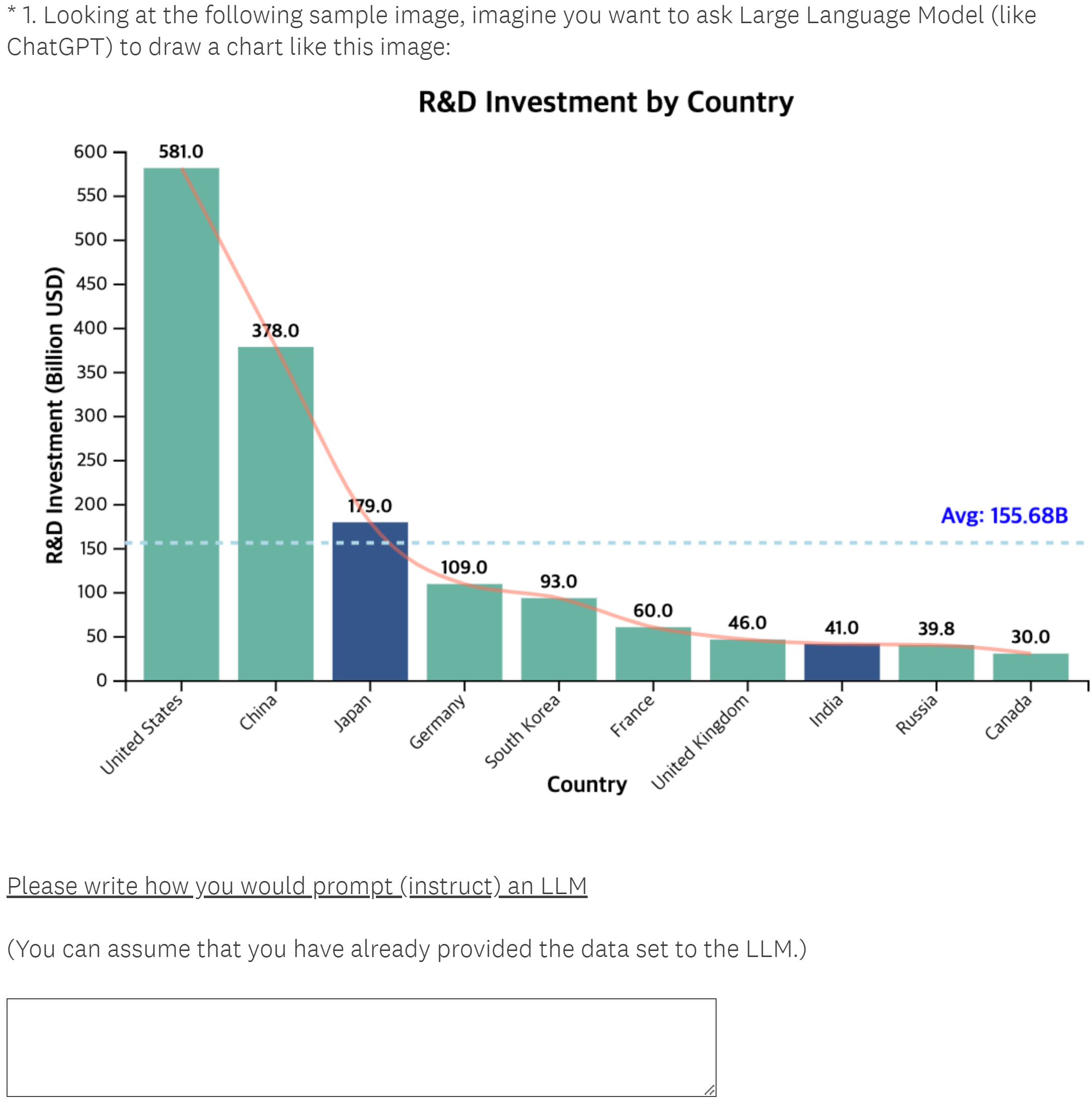}
    \captionsetup{justification=centering}
    \caption{ChartUIE-8K Human Study (Part 1) Example}
    \label{fig:uie_1}
\end{figure*}

\begin{figure*} [h]
    \centering
\includegraphics[height=0.9\textheight]{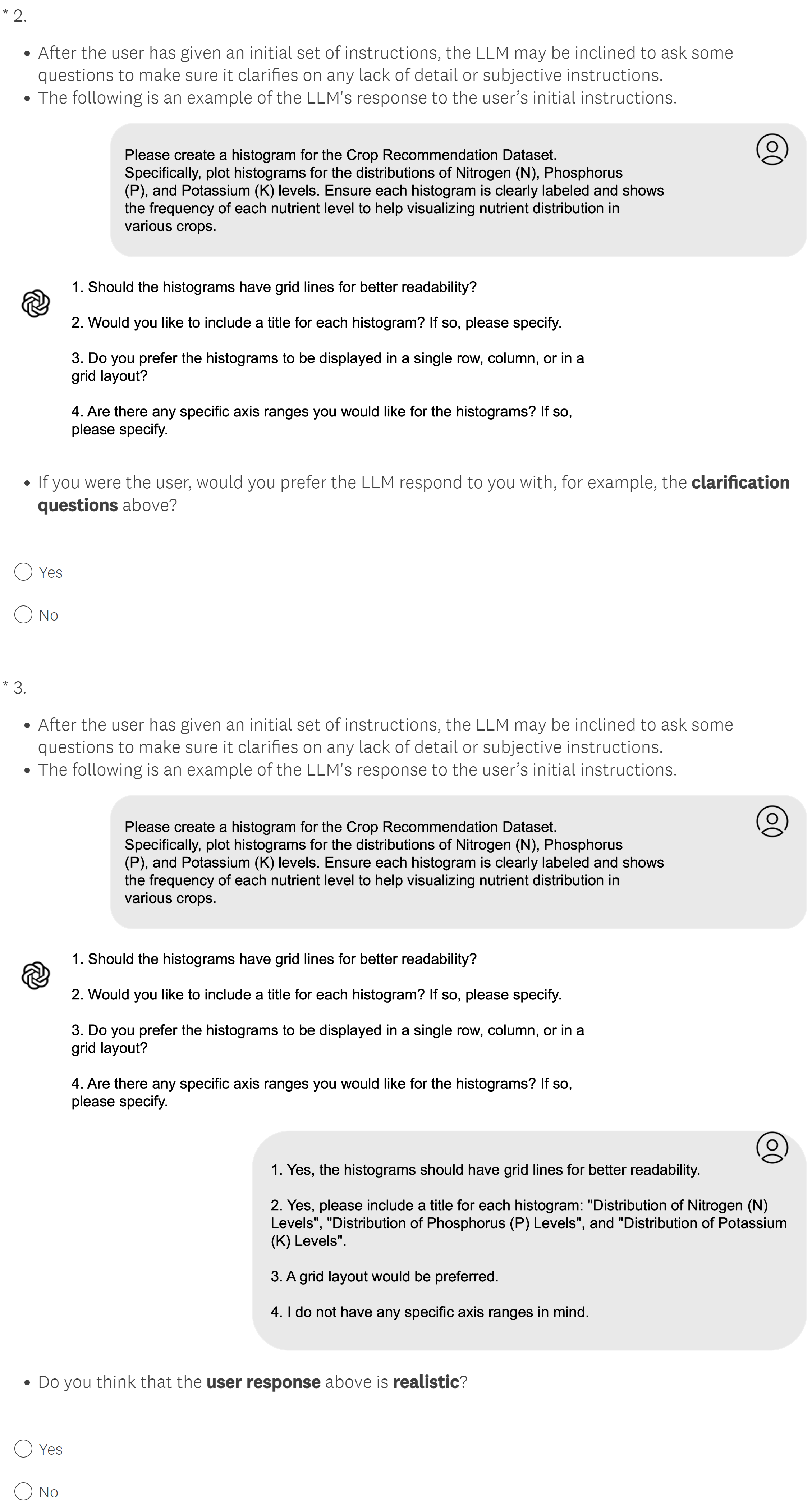}
    \captionsetup{justification=centering}
    \caption{ChartUIE-8K Human Study (Part 2) Example}
    \label{fig:uie_2}
\end{figure*}

We provide the precise instructions provided to the surveyors below. We further provide screenshots in Fig. \ref{fig:uie_1} and \ref{fig:uie_2}.

\begin{tcolorbox}[enhanced jigsaw,breakable,pad at break*=1mm,colback=darkred!5!white,colframe=darkred!50!black, colbacktitle=darkred!75!black,title=\textsc{ChartUIE-8K} Human Study (Part 1)]
\begin{itemize}
    \item Looking at the following sample image, imagine you want to ask Large Language Model (like ChatGPT) to draw a chart like this image:
\end{itemize}
\textit{[insert example chart image]}
\begin{itemize}
    \item \underline{Please write how you would prompt} \underline{(instruct) an LLM}
    \item (You can assume that you have already provided the dataset to the LLM.)
\end{itemize}
\textit{[insert text form field]}



\end{tcolorbox}

\begin{tcolorbox}[enhanced jigsaw,breakable,pad at break*=0.5mm,before skip=5pt,after skip=5pt,colback=darkred!5!white,colframe=darkred!50!black, colbacktitle=darkred!75!black,title=\textsc{ChartUIE-8K} Human Study (Part 2)]
\begin{itemize}[itemsep=1pt, parsep=0pt]
    \item After the user has given an initial set of instructions, the LLM may be inclined to ask some questions to make sure it clarifies on any lack of detail or subjective instructions.
   \item The following is an example of the LLM's response to the user’s initial instructions.
\end{itemize}
\textit{[insert sample initial instruction and questions]}

\begin{itemize}[itemsep=1pt, parsep=0pt]
    \item If you were the user, would you prefer the LLM respond to you with, for example, the \textbf{clarification questions} above?
\end{itemize}
\textit{[insert binary option (Yes or No)]}

\begin{itemize}[itemsep=1pt, parsep=0pt]
    \item After the user has given an initial set of instructions, the LLM may be inclined to ask some questions to make sure it clarifies on any lack of detail or subjective instructions.
   \item The following is an example of the LLM's response to the user’s initial instructions.
\end{itemize}
\textit{[insert sample initial instruction, questions, and answers for the questions]}

\begin{itemize}[itemsep=1pt, parsep=0pt]
    \item Do you think that the \textbf{user response} above is \textbf{realistic}?
\end{itemize}

\textit{[insert binary option (Yes or No)]}
\end{tcolorbox}

\clearpage 

\clearpage 

\clearpage 

\clearpage 



\section{LLM Configurations}\label{app:config}
Table~\ref{tab:llm_config2} shows the LLM  configurations we used. These configurations are determined \textit{a priori} to all experiments, and never changed.

\begin{table}[!h]
\centering
\resizebox{0.5\columnwidth}{!}{
\begin{tabular}{|c|c|c|c|}
\hline
\textbf{} & \textbf{Model} & \textbf{Temp.} & \textbf{Max Tokens} \\ \hline
\textbf{GPT 4o} & gpt-4o-2024-08-06 & 1 & 4096 \\
\textbf{Claude 3.5 Sonnet} & claude-3-5-sonnet-20240620 & 1 & 4096 \\
\textbf{Llama 3.1 70B} & llama-3.1-70b-versatile & 1 & 4096 \\
\textbf{Gemma 2 27B} & gemma-2-27b-it & 1 & 2048  \\ \hline
\end{tabular}
}
\caption{LLM Configurations}
\label{tab:llm_config2}
\end{table}

\section{Code Error Rate}\label{app:bench}
\begin{figure} [h]
    \centering
\includegraphics[width=0.5\linewidth]{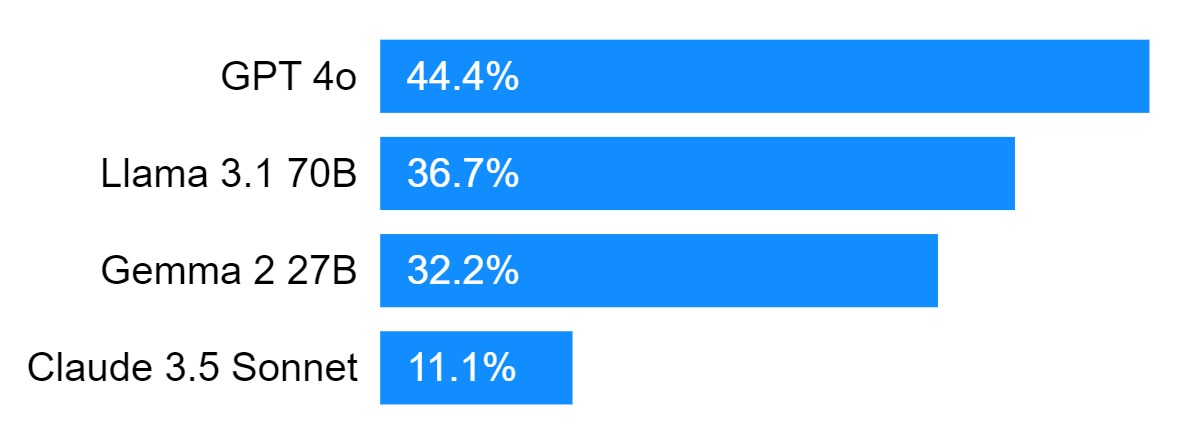}
    \caption{Execution error rates of the tested LLMs}
    \label{fig:error}
\end{figure}

Fig. \ref{fig:error} shows the mean execution error rates of the four tested LLMs. This is an inherent limitation of the original chart-generating LLM and does not relate to our method.

\section{Icon Attribution}

All icons are from \href{www.flaticon.com}{flaticon.com} except Retain(cycle-100) is from \href{https://icons8.com/}{icons8.com}. The icons and their respective authors are in Table \ref{table:icons}. The license to these icons belongs to their respective owners. We use them for non-profit academic use, with attribution as requested by the distributors.
\begin{table}[t]
\renewcommand{\arraystretch}{1.0}
\centering
\scriptsize
\resizebox{0.5\columnwidth}{!}{%
\begin{tabular}{|l|l|}
\hline
\textbf{Category (Icon)} & \textbf{Author(s)} \\ \hline
Retain (cycle-100) & Icons8 \\ \hline
Discard (remove) & Pixel perfect \\ \hline
Add (add) & Arkinasi \\ \hline
Task (checklist) & juicy\_fish \\ \hline
Purpose (target) & Freepik \\ \hline
Audience (users-avatar) & SeyDesigner \\ \hline
Basic Criteria (list) & Smashicons \\ \hline
Addition (add) & Digby Garrett \\ \hline
Query (question) & Md Tanvirul Haque \\ \hline
Binarize (file-management) & iconsmind \\ \hline
Social Science (network) & Freepik \\ \hline
Biology (book) & Freepik \\ \hline
Box Chart (chart) & Freepik \\ \hline
Stacked Chart (column-chart) & Freepik \\ \hline
LLM (big-data) & LAFS \\ \hline
AI (ai) & Freepik \\ \hline
Number 1 (circle-1) & iconographics \\ \hline
Number 2 (circle-2) & iconographics \\ \hline
Number 3 (number-3) & BS Editing \\ \hline
LLM to Evaluate (dashboard) & xnimrodx \\ \hline
Data Set Filter (database-structure) & Uniconlabs \\ \hline
Chart Annotations (notes) & Freepik \\ \hline
Data Set Crawl (bug) & Freepik \\ \hline
Coding (code-review) & Freepik \\ \hline
Evaluate (criteria) & cah nggunug \\ \hline
Edit (modifed) & Uniconlabs \\ \hline
\end{tabular}%
}
\caption{Icons and Respective Authors}
\label{table:icons}
\end{table}

\end{document}